\documentclass{article}
\pdfoutput=1
\usepackage{times}  
\usepackage{helvet} 
\usepackage{courier}  
\usepackage[hyphens]{url}  
\usepackage{graphicx} 
\usepackage{graphicx}  
 \newlength\titlebox 
\title{Proofs and additional experiments on \protect\textit{Second order techniques for learning time-series with structural breaks}}
\author{
  Takayuki Osogami\\
  IBM Research - Tokyo\\
  {\tt osogami@jp.ibm.com}
}

\usepackage{natbib}
\usepackage{amsmath}
\usepackage{amsthm}
\usepackage{amssymb}
\usepackage{bm}
\usepackage{algorithm, algpseudocode}

\newcommand{\bfc}{{\mathbf{c}}}

\newcommand{\bfg}{{\mathbf{g}}}
\newcommand{\bfh}{{\mathbf{h}}}

\newcommand{\bfk}{{\mathbf{k}}}

\newcommand{\bfu}{{\mathbf{u}}}

\newcommand{\bfx}{{\mathbf{x}}}

\newcommand{\bfH}{{\mathbf{H}}}
\newcommand{\bfI}{{\mathbf{I}}}

\newcommand{\bfM}{{\mathbf{M}}}

\newcommand{\bfO}{{\mathbf{O}}}

\newcommand{\hatbfx}{{\mathbf{\hat x}}}

\newcommand{\hatbfH}{{\mathbf{\hat H}}}

\newcommand{\checkbfx}{{\mathbf{\check x}}}

\newcommand{\checkbfH}{{\mathbf{\check H}}}

\newcommand{\tildebfH}{{\mathbf{\tilde H}}}

\newcommand{\bstheta}{{\boldsymbol{\theta}}}

\newcommand{\tildebstheta}{{\boldsymbol{\tilde\theta}}}

\newcommand{\hatbstheta}{{\boldsymbol{\hat\theta}}}

\newcommand{\checkbstheta}{{\boldsymbol{\check\theta}}}

\begin{document}
\maketitle

\begin{abstract}
  We provide complete proofs of the lemmas about the properties of the
  regularized loss function that is used in the second order
  techniques for learning time-series with structural breaks in
  \citet{AAAI21}.  In addition, we show experimental results that
  support the validity of the techniques.
\end{abstract}

\appendix
\section{Introduction}

We study a nonstationary time-series model, $f_{\bstheta_t}(\cdot)$,
having time-varying weights, $\bstheta_t$ at step $t$.  Given an input
$\bfx_t$ at $t$, the model makes a prediction $\hat y_t \equiv
f_{\bstheta_t}(\bfx_t)$ about the target $y_t$.  In online learning,
we update $\bstheta_t$ to $\bstheta_{t+1}$ after observing $y_t$ and
use $\bstheta_{t+1}$ to make the prediction, $\hat y_{t+1}\equiv
f_{\bstheta_{t+1}}(\bfx_{t+1})$, about the next target $y_{t+1}$.

\citet{AAAI21} proposes to find the weights that, at every step $t$,
minimize the following weighted mean squared error with a
regularization term:
\setcounter{equation}{6}
\begin{align}
  \tilde L_t(\bstheta)
  & = L_t(\bstheta) + (\lambda/2) \, ||\hatbstheta||^2_{\bfH_t},
\end{align}
where
\setcounter{equation}{0}
\begin{align}
  L_t(\bstheta)
  & \equiv \frac{1}{2} \,
  \sum_{d=0}^{t-1} \gamma^d \, \left(f_{\bstheta}(\bfx_{t-d}) - y_{t-d} \right)^2,
\end{align}
and $\bfH_t$ is the Hessian of \eqref{eq:WMSE}:
\setcounter{equation}{2}
\begin{align}
  \bfH_t
  & \equiv \sum_{d=0}^{t-1} \gamma^d \, \bfx_{t-d} \, \bfx_{t-d}^\top
  = \gamma \, \bfH_{t-1} + \bfx_t \, \bfx_t^\top.
\end{align}

\citet{AAAI21} presents the following lemmas about the properties of
the regularized loss function \eqref{eq:reg_loss}.

\paragraph{Lemma~\ref{lemma:reg} \citep{AAAI21}}
{\it
  For linear models, the minimizer of the regularized loss function
  \eqref{eq:reg_loss} is given by
  \begin{align}
    \tildebstheta^\star = \tildebfH_t^{-1} \, \bfg_t,
    \notag
  \end{align}
  where
  \begin{align}
    \tildebfH_t
    = \gamma \, \tildebfH_{t-1}
    + \bfx_t \, \bfx_t^\top
    + \lambda \, \hatbfx_t \, \hatbfx_t^\top
    \notag
  \end{align}
    and $\tildebfH_0=\bfO$.    
  Then $\tildebfH_t^{-1}$ can be computed
  from $\bfH_{t-1}^{-1}$ in $O(n^2)$ time by applying the
  Sherman-Morrison lemma twice.
}

\paragraph{Lemma~\ref{lemma:invariance} \citep{AAAI21}}
{\it
  Consider an invertible linear transformation $\bfM$ of order $n-1$, and
  let
  \begin{align}
    \checkbfx_t' = \bfM \, \checkbfx_t
    \notag
  \end{align}
  for each $t$.  Let the
  weights except the intercept be contravariate to $\bfM$ in that
  $\bfM$ transforms $\checkbstheta^\top$ into
  \begin{align}
    \checkbstheta'^\top=\checkbstheta^\top\bfM^{-1}.
    \notag
  \end{align}
  Then the loss function
  \eqref{eq:reg_loss} is invariant to $\bfM$.
}

\paragraph{}
The rest of the article is organized as follows.  In
Section~\ref{sec:proofs}, we prove Lemma \ref{lemma:reg} and Lemma
\ref{lemma:invariance}.  In Section~\ref{sec:additional_exp}, we
provide the experimental results that were omitted in \citet{AAAI21}
due to space considerations.  Throughout we refer to the equations,
lemmas, and figures in \citet{AAAI21} with the same labels as
\citet{AAAI21}.  Specifically, Equations (1)-(12), Lemmas 1-3, and
Figures 1-2 refer to those appeared in \citet{AAAI21}.

\setcounter{equation}{12}
\setcounter{figure}{3}

\section{Proofs}
\label{sec:proofs}

\subsection{Proof of Lemma \ref{lemma:reg}}

We will use the following notations:
\begin{align}
  \bfH_t & = \left(\begin{array}{cc}
    h_t & \bfh_t^\top \\
    \bfh_t & \checkbfH_t
  \end{array}\right),
  \\
  \hatbfH_t & = \left(\begin{array}{cc}
    0 & \mathbf{0}^\top \\
    \mathbf{0} & \checkbfH_t
  \end{array}\right),
  \\
  \hatbfx_t & = \left(\begin{array}{c}
    0\\
    \checkbfx_t
  \end{array}\right),
\end{align}
where $\hatbfH_t$ can be written recursively as follows:
\begin{align}
  \hatbfH_t
  & = \sum_{d=0}^{t-1} \gamma^d \, \hatbfx_{t-d} \, \hatbfx_{t-d}^\top \\
  & = \gamma \, \hatbfH_{t-1} + \hatbfx_{t} \, \hatbfx_{t}^\top.
  \label{eq:rec_hatH}
\end{align}
We can then write our loss function as follows:
\begin{align}
  \tilde L_t(\bstheta)
  & = L_t(\bstheta) + \frac{\lambda}{2} \, \bstheta^\top \, \hatbfH_t \, \bstheta.
\end{align}
Because $\tilde L_t(\cdot)$ is a quadratic function, its minimizer is
given by the $\tildebstheta^\star$ in the lemma, where
\begin{align}
  \tildebfH_t \equiv \bfH_t + \lambda \, \hatbfH_t.
\end{align}
By \eqref{eq:rec_H} and \eqref{eq:rec_hatH}, we can write
$\tildebfH_t$ recursively as
\begin{align}
  \tildebfH_t
  & = \gamma \, \tildebfH_{t-1}
  + \bfx_t \, \bfx_t^\top
  + \lambda \, \hatbfx_t \, \hatbfx_t^\top.
  \label{eq:rec_tildeH}
\end{align}
This completes the proof of the lemma.

\subsection{Proof of Lemma \ref{lemma:invariance}}

It is known that $L_t(\cdot)$ is invariant to $\bfM$.  Specifically,
  $\bfM$ transforms $L_t(\cdot)$ into $L_t'(\cdot)$, where
  \begin{align}
    L_t'(\bstheta')
    & = \frac{1}{2} \sum_{d=0}^{t-1} \gamma^d \, \left(
    \theta^{(0)} + \checkbstheta'^\top \checkbfx_t' - y_t
    \right) \\
    & = \frac{1}{2} \sum_{d=0}^{t-1} \gamma^d \, \left(
    \theta^{(0)} + \checkbstheta^\top \checkbfx_t - y_t
    \right) \\ 
    & = L_t(\bstheta).
  \end{align}
  
  It thus suffices to show that the regularization term is invariant
  to $\bfM$.  Observe that $\bfM$ transforms
  $||\hatbstheta||_{\bfH_t}^2$ into
    $||\hatbstheta'||^2_{\bfH'_t}
    = \checkbstheta'^\top \checkbfH'_t \checkbstheta'$,
where
  \begin{align}
    \checkbfH'_t
    & = \sum_{d=0}^{t-1} \gamma^d \, \checkbfx'_{t-d} \, \checkbfx'^\top_{t-d} \\
    & = \sum_{d=0}^{t-1} \gamma^d \, \bfM \, \checkbfx_{t-d} \, \checkbfx^\top_{t-d} \, \bfM^\top \\
    & = \bfM \, \checkbfH_t \, \bfM^\top.
  \end{align}
Thus,
$||\hatbstheta'||^2_{\bfH'_t}
    = \checkbstheta^\top \, \checkbfH_t \, \checkbstheta
    = ||\hatbstheta||^2_{\bfH_t}$,
proving the lemma.

\section{Additional experiments}
\label{sec:additional_exp}

\subsection{Experiments on regularization}

\subsubsection{Sensitivity of L2 regularization to the transformation of the coordinates}

\begin{figure}[t]
  \centering
  \includegraphics[width=0.5\linewidth]{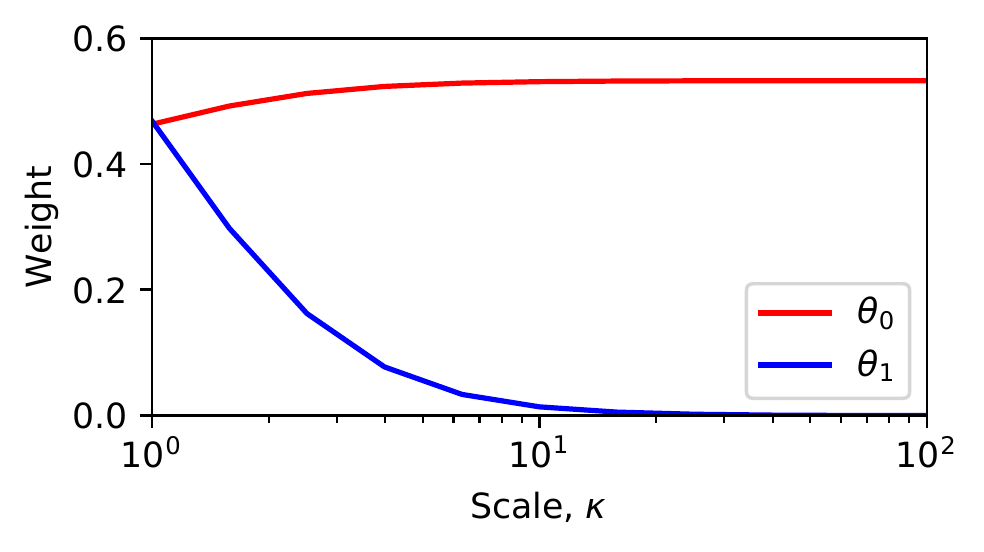}
  \caption{The weight $\bstheta=(\theta_0, \theta_1)$ given by ridge
    regression (L2 regularization with $\lambda=1$), where a variable
    $x_1$ is scaled as $\hat x_1=x_1/\kappa$, and the corresponding
    weight $\hat\theta_1$ is unscaled as
    $\theta_1=\hat\theta_1/\kappa$ ({\it i.e.},
    $y=\theta_0x_0+\theta_1x_1=\theta_0x_0+\hat\theta_1\hat x_1$).
    Training data is generated in a way that each explanatory
    variable, $x_i$ for $i\in\{0,1\}$, is i.i.d.\ with the standard
    normal distribution, and the target variable is
    $y=x_0+x_1+\varepsilon$, where noise $\varepsilon$ is i.i.d.\ with
    the standard normal distribution.}
  \label{fig:weight}
\end{figure}

In Figure~\ref{fig:weight}, we show that standard L2 regularization is
sensitive to the transformation of the coordinates of explanatory
variables.  Recall that one minimizes the following loss function with
standard L2 regularization:
\setcounter{equation}{4}
\begin{align}
  \dot L_t(\bstheta)
  & = L_t(\bstheta) + (\lambda/2) \, ||\bstheta||^2_2.
\end{align}

Here, we have two explanatory variables, $x_1$ and $x_2$, and the
training data is generated according to a linear model
$y=\theta_1x_1+\theta_2x_2+\varepsilon$, where $\theta_1=\theta_2=1$.
In online learning, we often cannot normalize the variables (to have
unit variance) a priori.  If a scaled variable $\hat x_1=x_1/\kappa$
is observed (and not normalized), the corresponding true weight is
also scaled $\hat\theta_1=\kappa\theta_1$.  Depending on the value of
$\kappa$, L2 regularization has varying effect on $\hat\theta_1$.  For
a large $\kappa$, the magnitude of the estimated $\hat\theta_1$ is
large and hence is reduced by a large amount by L2 regularization.
This however implies that the corresponding unscaled weight
$\theta_1=\hat\theta_1/\kappa$ gets smaller than what is given when
the variables are normalized (to have unit variance).

\subsubsection{Effectiveness of the proposed regularization}

\begin{figure*}[tb]
  \centering
  \begin{minipage}{0.32\linewidth}
    \centering
    \includegraphics[width=\linewidth]{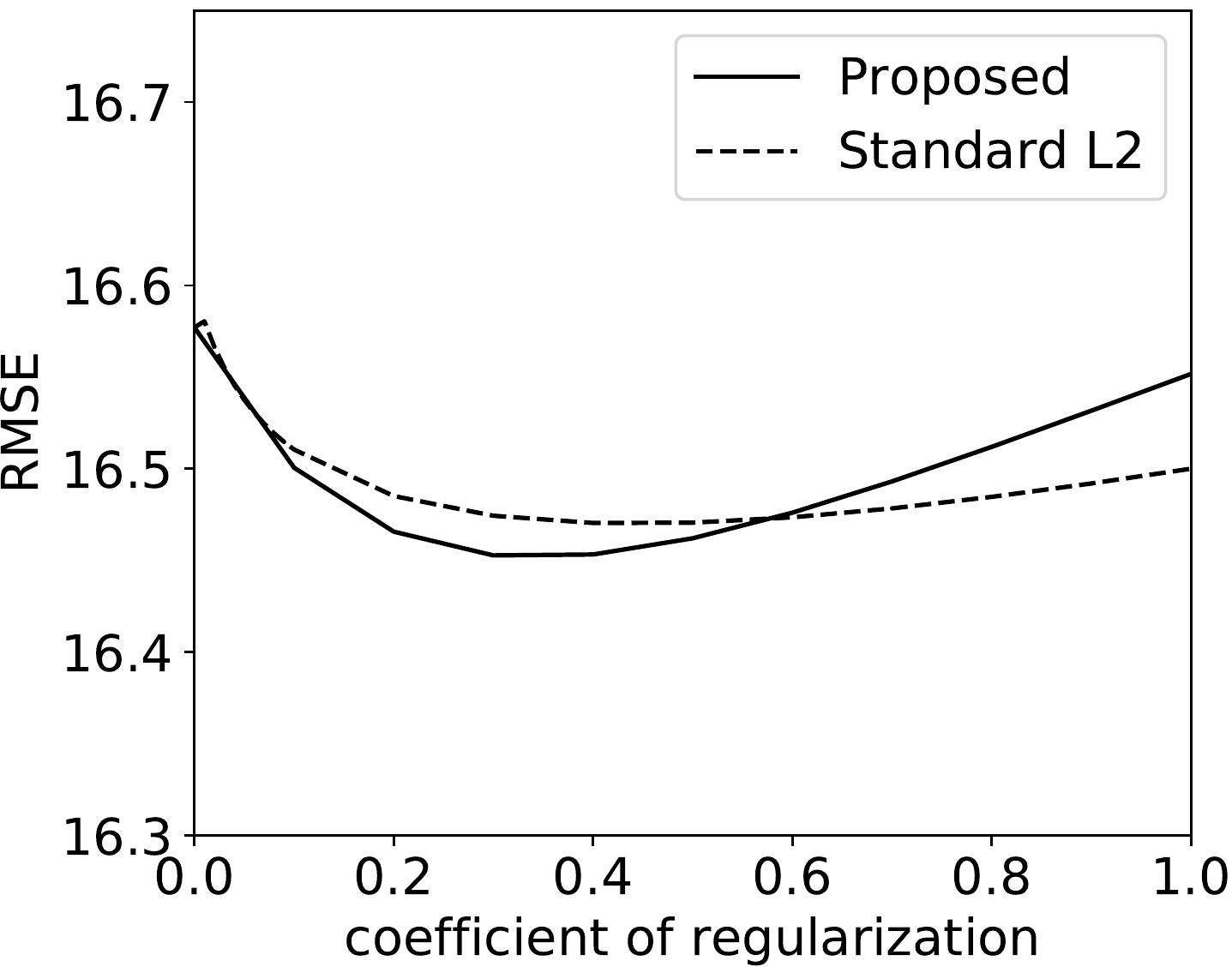}
    (a) Order 2
  \end{minipage}
  \begin{minipage}{0.32\linewidth}
    \centering
    \includegraphics[width=\linewidth]{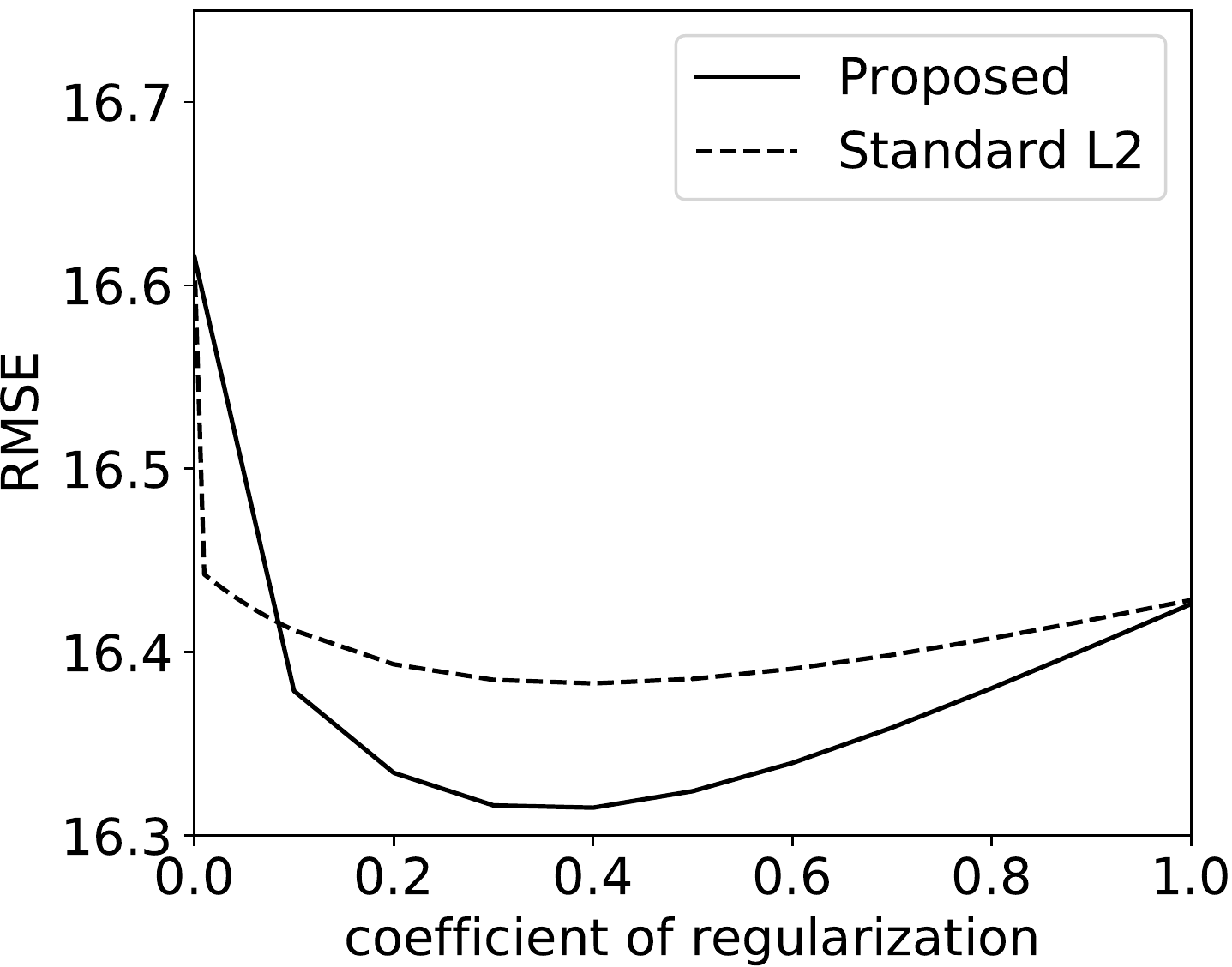}
    (b) Order 3
  \end{minipage}
  \begin{minipage}{0.32\linewidth}
    \centering
    \includegraphics[width=\linewidth]{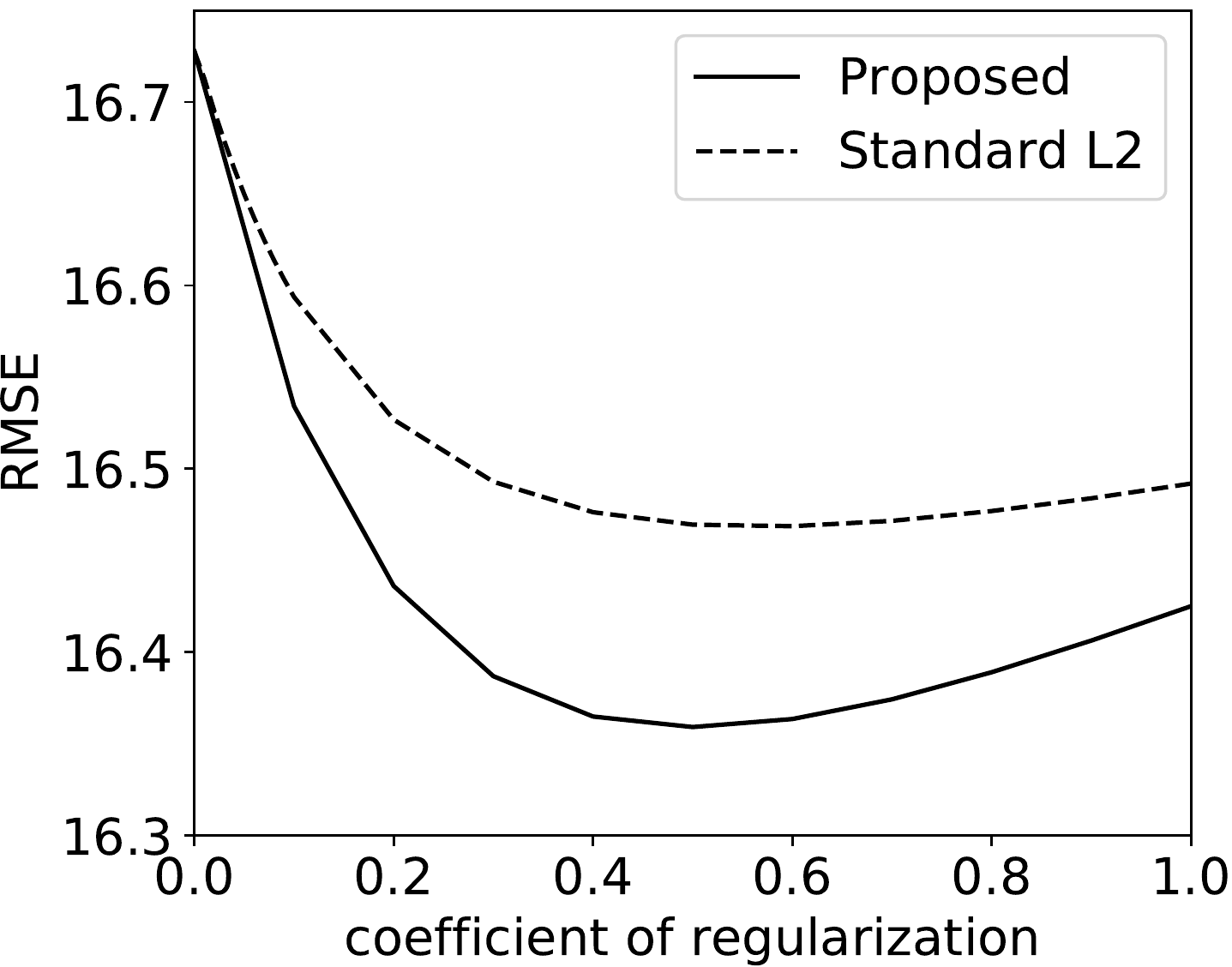}
    (c) Order 4
  \end{minipage}
  \caption{The RMSE of predictions by the AR model of varying order
    with the optimal parameters that minimize the loss function with
    L2 or our regularization, where the coefficient of regularization
    $\lambda$ (for L2, $\lambda/100$) is varied.}
  \label{fig:L2}
\end{figure*}

In Figure~\ref{fig:L2}, we compare the effectiveness of our
regularization against L2 regularization.  Here, we learn a
time-series of the monthly sunspot number from January 1749 to
December 1983 (2,820
steps)\footnote{https://datamarket.com/data/set/22t4/}.  We use this
dataset primarily because it exhibits large fluctuations of the
magnitude.  We train autoregressive (AR) models of varying order, as
indicated in each panel of the figure.  At each step, the parameters
are optimized in a way that they minimize either the loss function
with L2 regularization \eqref{eq:loss_standardL2} or the one with our
regularization \eqref{eq:loss_invariant}, where we fix $\gamma=0.99$.

Overall, our regularization compares favorably against L2
regularization.  In particular, the best RMSE of 16.31 is achieved by
our regularization at $\lambda=0.4$ for the model with the third order
(Figure~\ref{fig:L2}~(b)), while L2 regularization cannot reduce the
RMSE below 16.38 for any $\lambda$ and for any order.  Although the
effectiveness of regularization depends on particular data, the
results of this experiment suggest that our regularization not only
can be performed in $O(n^2)$ time but also has the expected effect of
regularization, sometimes outperforming L2 regularization ({\it e.g.}
when a time-series involves large fluctuations).

\subsection{Experiments on recursively computing pseudo-inverse}

\citet{AAAI21} proposes to update the inverse Hessian based on the following lemma:
\paragraph{Lemma~\ref{lemma:update_pseudo} \citep{AAAI21}}
{\it
  For a symmetric $\bfH \in \mathbb{R}^{n\times n}$ and
  $\bfc\in\mathbb{R}^n$, let
  \begin{align}
    \bfu   & \equiv (\bfI - \bfH^+ \bfH) \, \bfc, \notag\\
    \bfu^+ & \equiv \bfu/(\bfu^\top\bfc), \mbox{ and } \notag\\
    \bfk   & \equiv \bfH^+ \bfc.
  \end{align}
  Then the pseudo-inverse $(\bfH+\bfc\,\bfc^\top)^+$
  can be computed from $\bfH^+$ as follows:
  if $\bfu^\top\bfc > 0$, then
  \begin{align}
    (\bfH+\bfc\,\bfc^\top)^+
    & =
    \bfH^+ - \bfk \, (\bfu^+)^\top - \bfu^+ \bfk^\top
    + (1 + \bfc^\top \bfk) \, \bfu^+ (\bfu^+)^\top;
    \notag
  \end{align}
  if $\bfu^\top\bfc = 0$, then
  \begin{align}
    (\bfH+\bfc\,\bfc^\top)^+
    & =
    \bfH^+ - \bfk \, \bfk^\top / (1+\bfc^\top \bfH^+ \bfc).
    \notag
  \end{align}
}

\begin{figure*}[t]
  \begin{minipage}{0.33\linewidth}
    \centering
    \includegraphics[width=\linewidth]{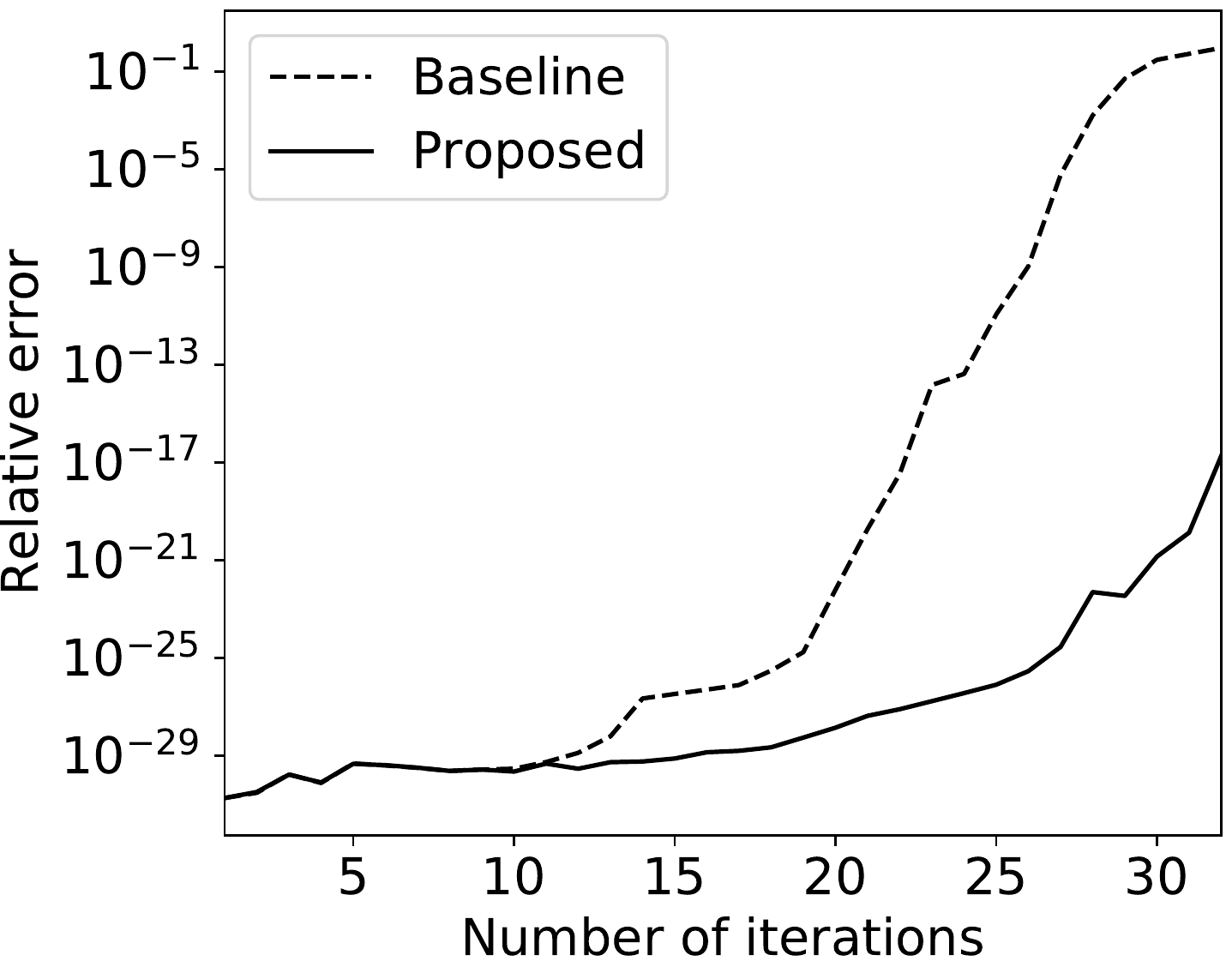}\\
    (a) $n=32$
  \end{minipage}
  \begin{minipage}{0.33\linewidth}
    \centering
    \includegraphics[width=\linewidth]{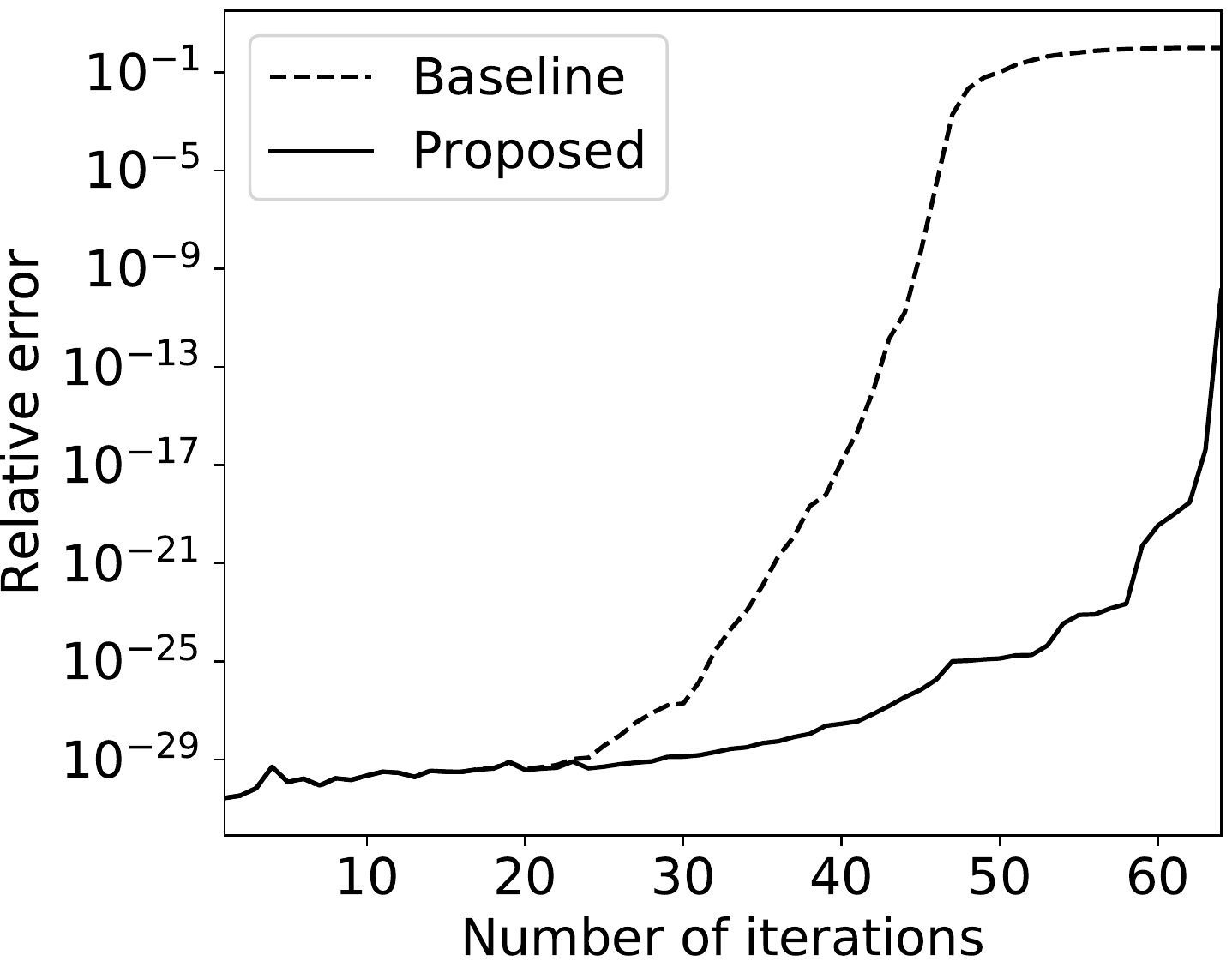}\\
    (b) $n=64$
  \end{minipage}
  \begin{minipage}{0.33\linewidth}
    \centering
    \includegraphics[width=\linewidth]{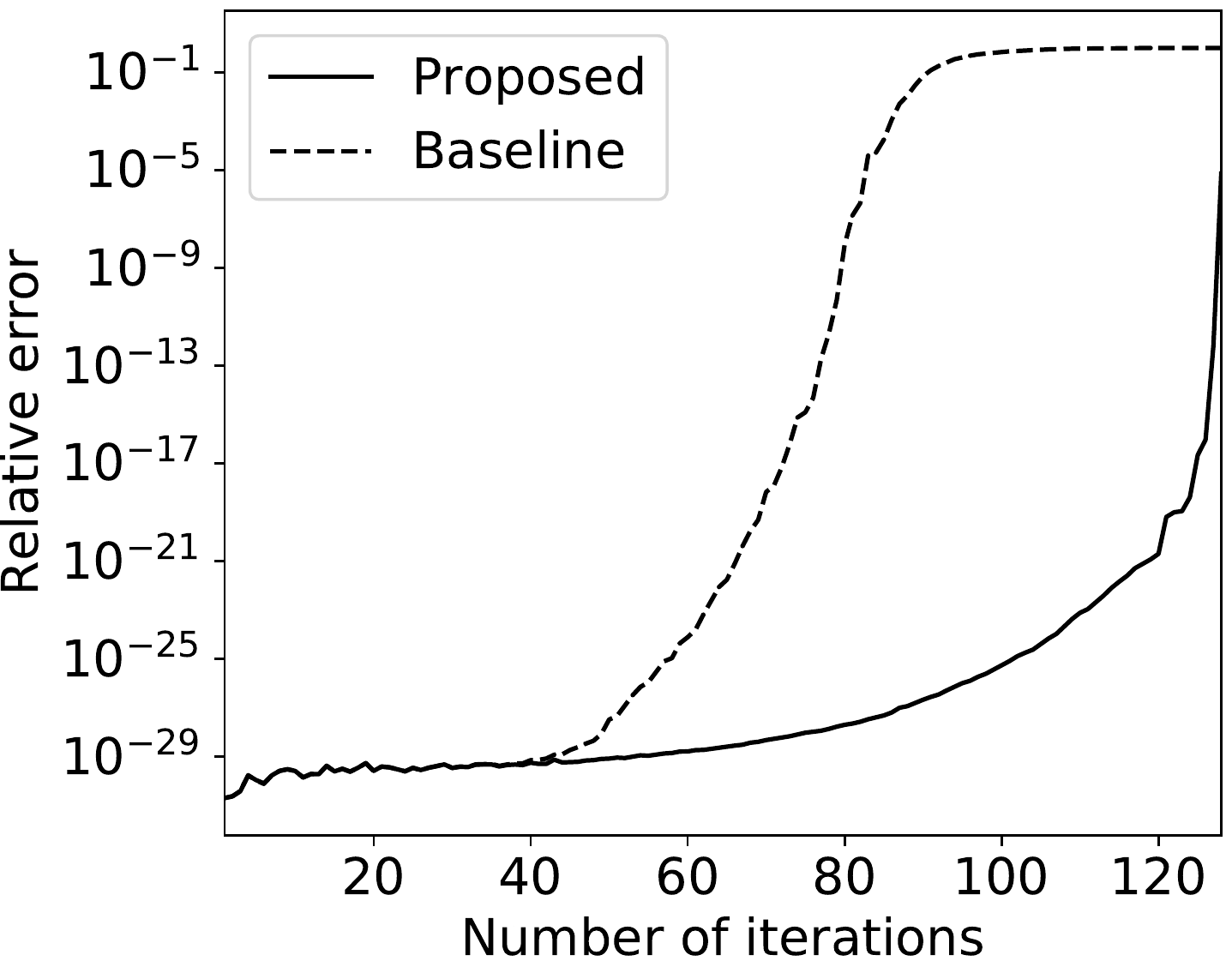}\\
    (c) $n=128$
  \end{minipage}
 \caption{Numerical error accumulated in recursively computed
   pseudo-inverse.}
  \label{fig:pinv}
\end{figure*}

\paragraph{}
Figure~\ref{fig:pinv} shows the numerical error accumulated in
recursively computed pseudo-inverse with two methods: Proposed and
Baseline.  Proposed is the one based on
Lemma~\ref{lemma:update_pseudo}.  Baseline differs from 
Proposed in the following two definitions: $\bfu \equiv (\bfI -
\bfH \, \bfH^+) \, \bfc$ and $\bfu^+\equiv \bfu/||\bfu||^2$.

Specifically, we recursively compute the pseudo-inverse of the
$n\times n$ matrix $\bfH_t = \bfH_{t-1} + \mathbf{x}_t
\,\mathbf{x}_t^\top$ for $t=1, \ldots, n$, where $\bfH_0=\bfO$, and
$\mathbf{x}_t$ is a column vector of length $n$, whose elements are
i.i.d.\ according to the standard
normal distribution.  We then evaluate the relative error of a
recursively compute matrix, which is the sum of the squared error of
each element divided by the sum of the squared value of each element
of the ground truth matrix, which is computed non-recursively.

Figure~\ref{fig:pinv} suggests that Proposed
is up to $10^{20}$ times more accurate than Baseline.

\subsection{Details of the results from the experiments with the synthetic time-series}

\begin{figure*}[th]
  \centering
  \includegraphics[width=0.5\linewidth]{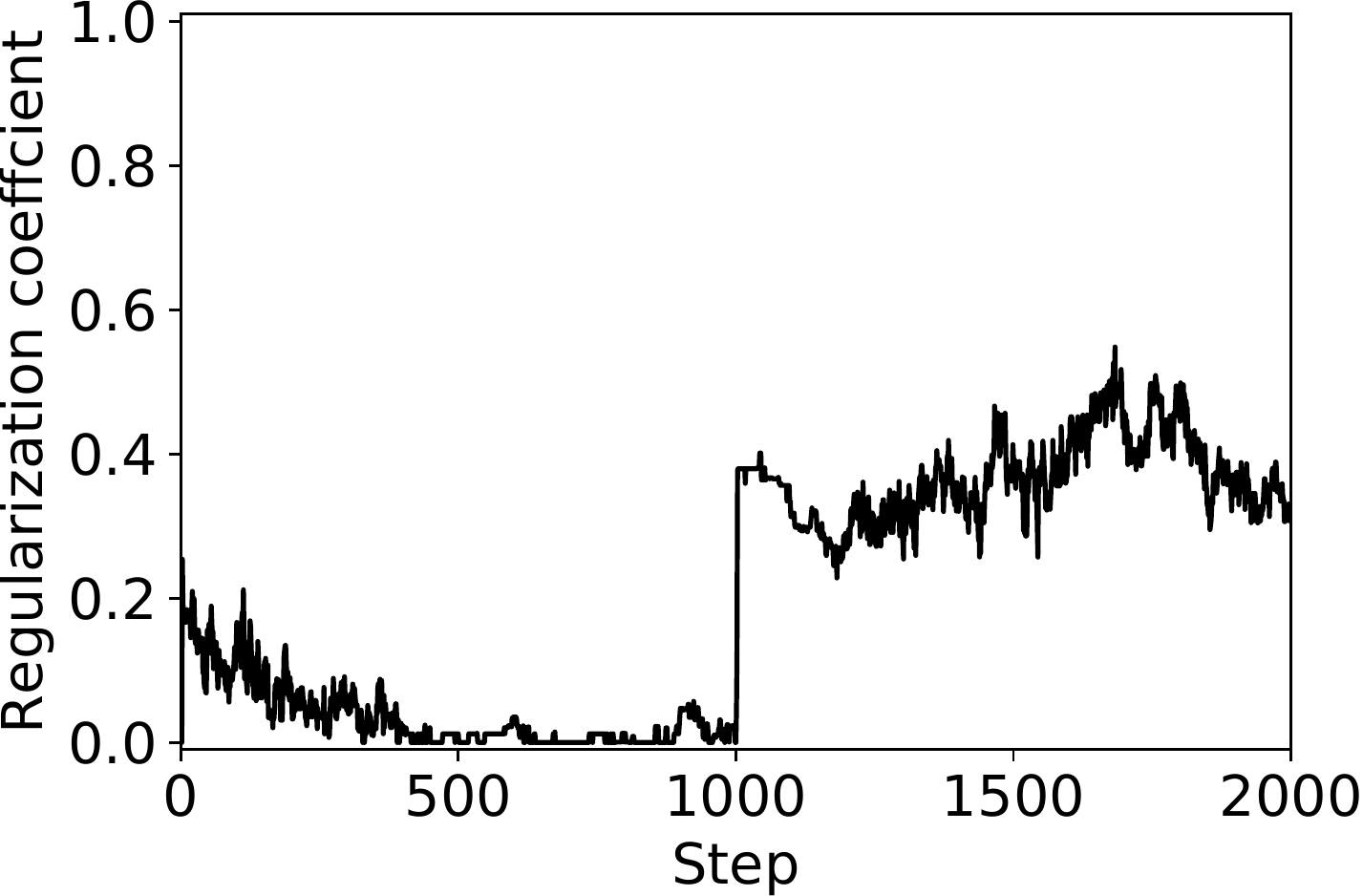}
  \caption{   
    The regularization-coefficient used by
    Algorithm~\ref{alg:adaptive} at each step in the experiments with
    the synthetic time-series.
  }
  \label{fig:synthetic2}
\end{figure*}

Figure~\ref{fig:synthetic2} shows the values of the regularization coefficient $\lambda$ used by
Algorithm~\ref{alg:adaptive} at each step in the experiments with the
synthetic time-series, where the corresponding value of the forgetting
rate $\gamma$ is shown in Figure~\ref{fig:synthetic}~(c).

\subsection{Details of the results from the experiments with stock indices}

\begin{figure*}[h]
  \begin{centering}
    \underline{\Large Detailed results on SPX}\\
    \ \\
  \end{centering}
  \includegraphics[width=0.33\linewidth]{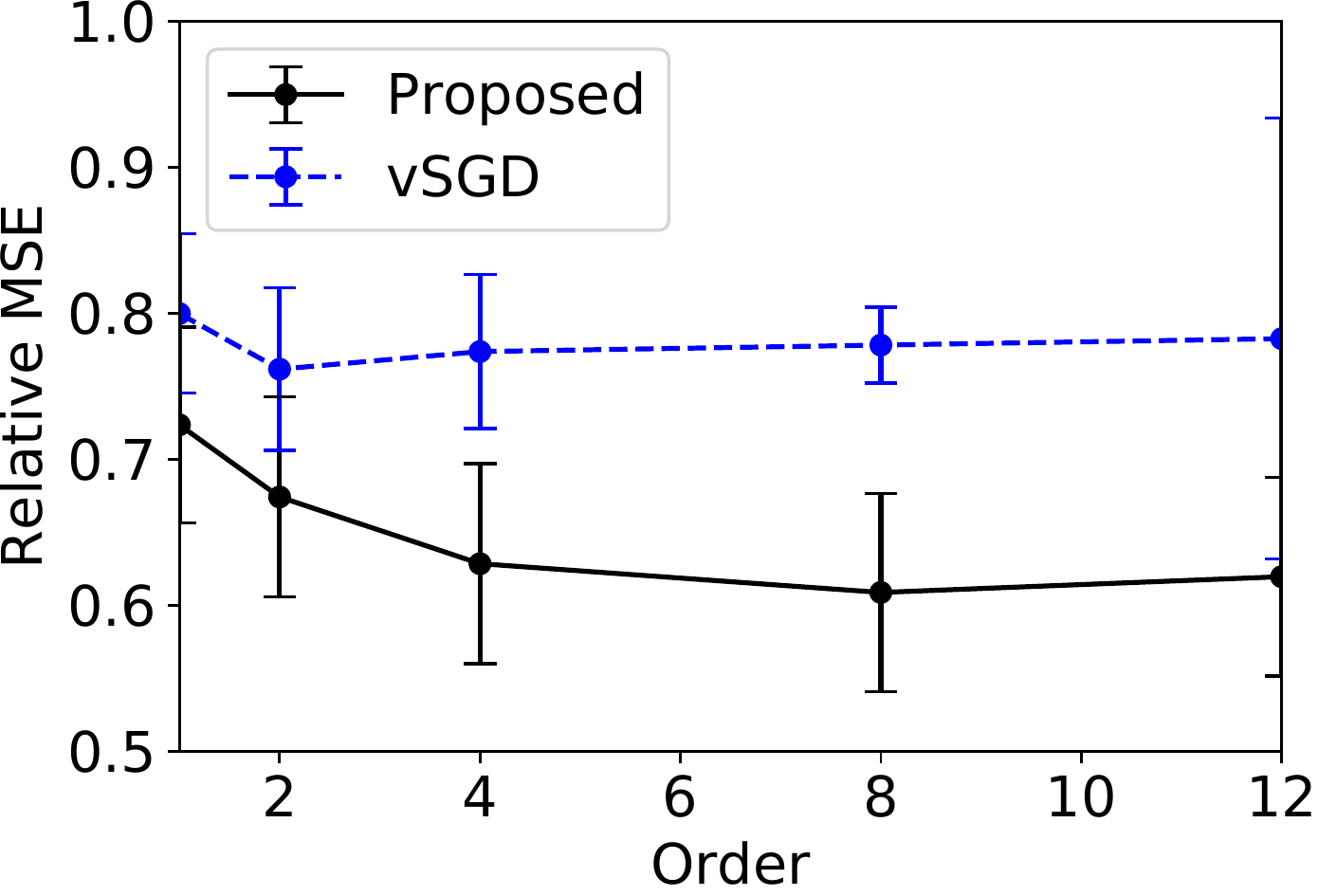}
  \includegraphics[width=0.33\linewidth]{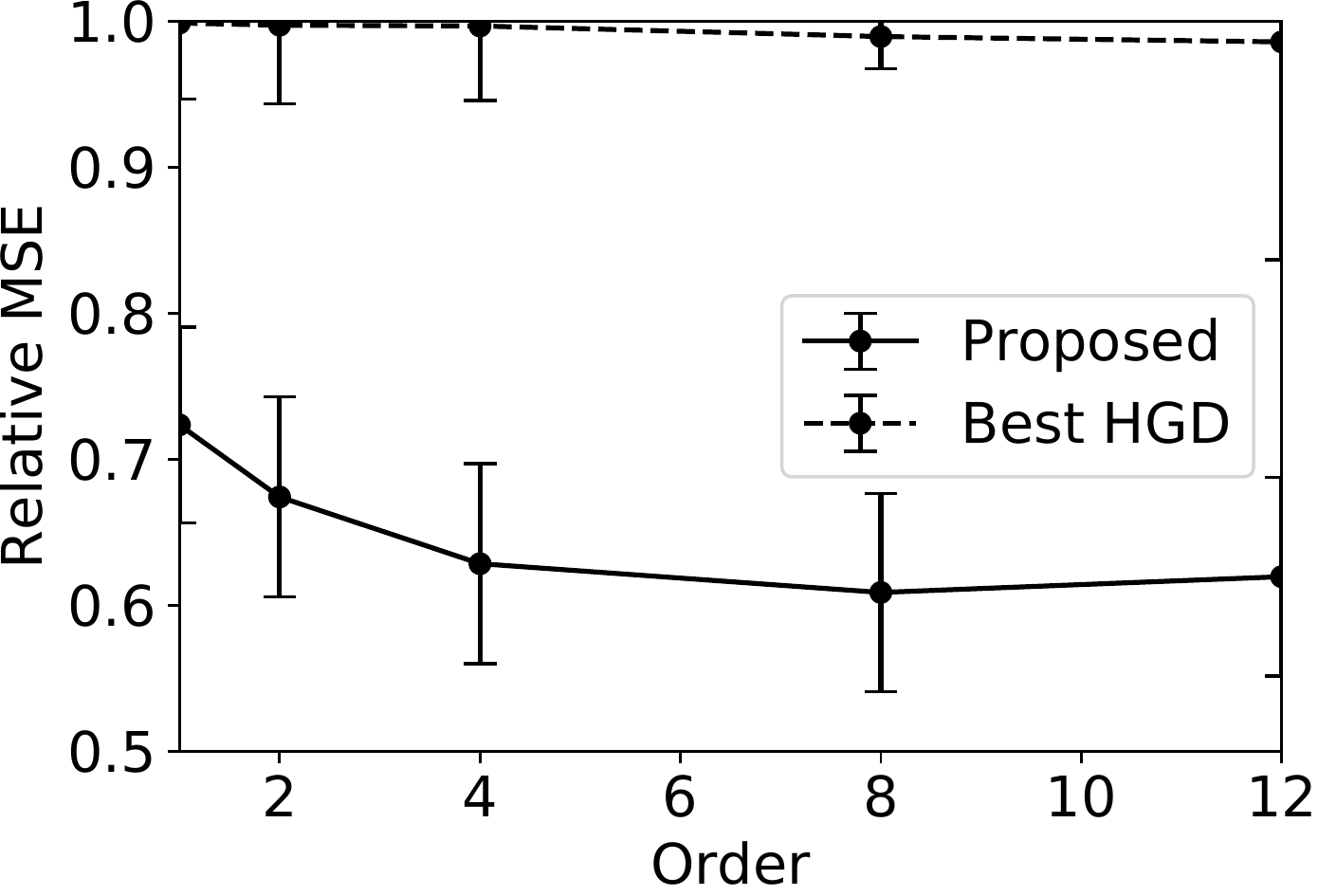}
  \includegraphics[width=0.33\linewidth]{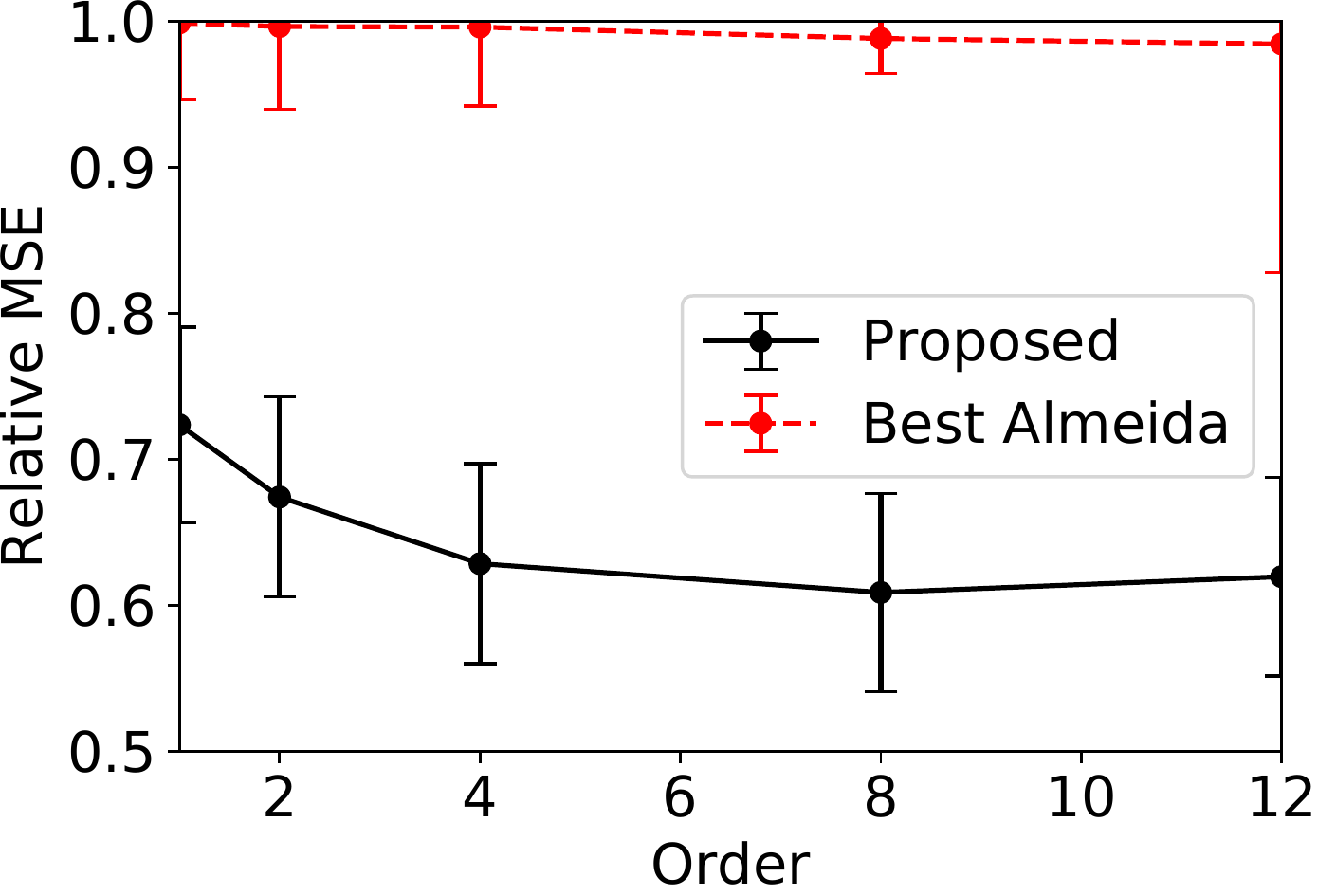}
  \includegraphics[width=0.33\linewidth]{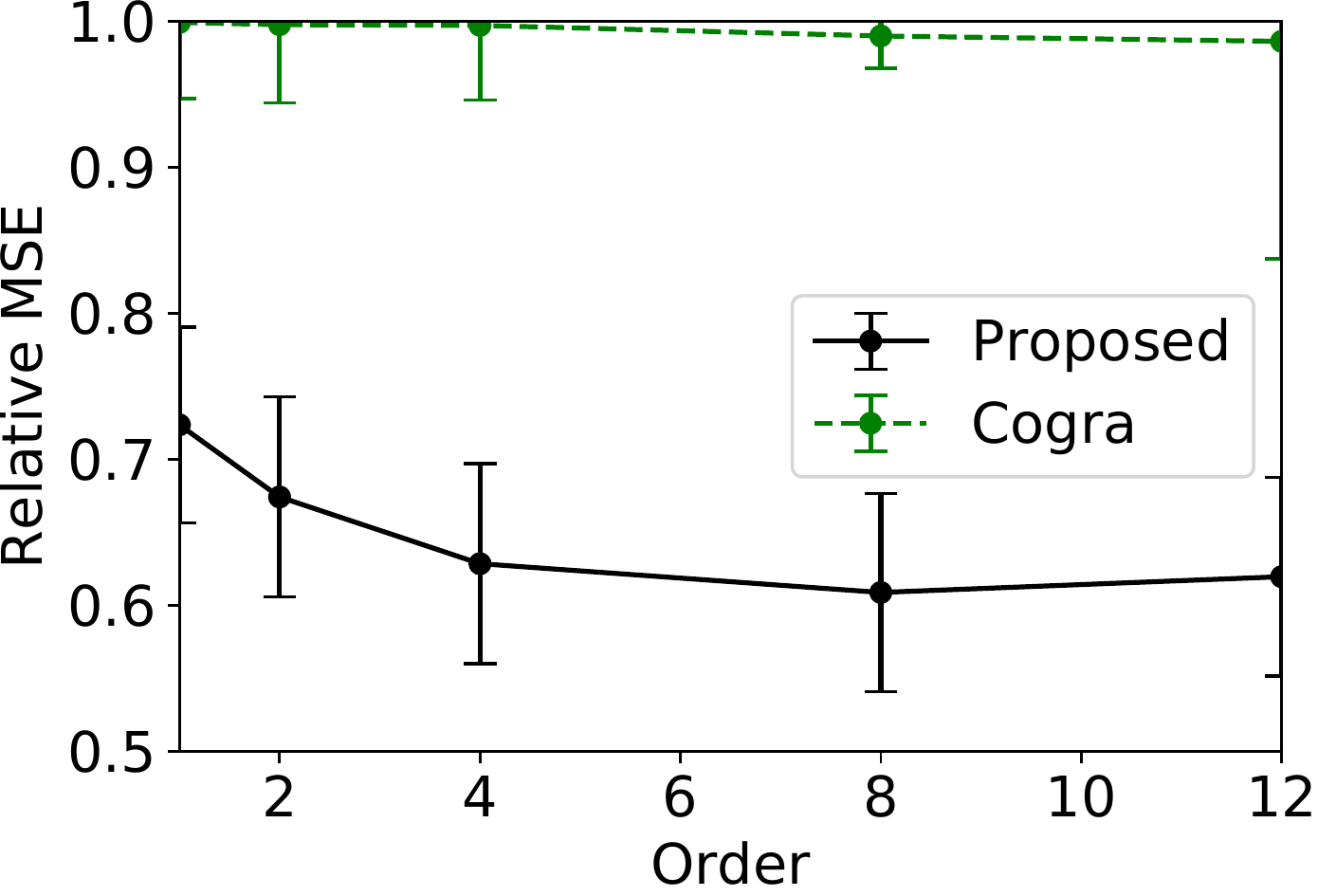}
  \includegraphics[width=0.33\linewidth]{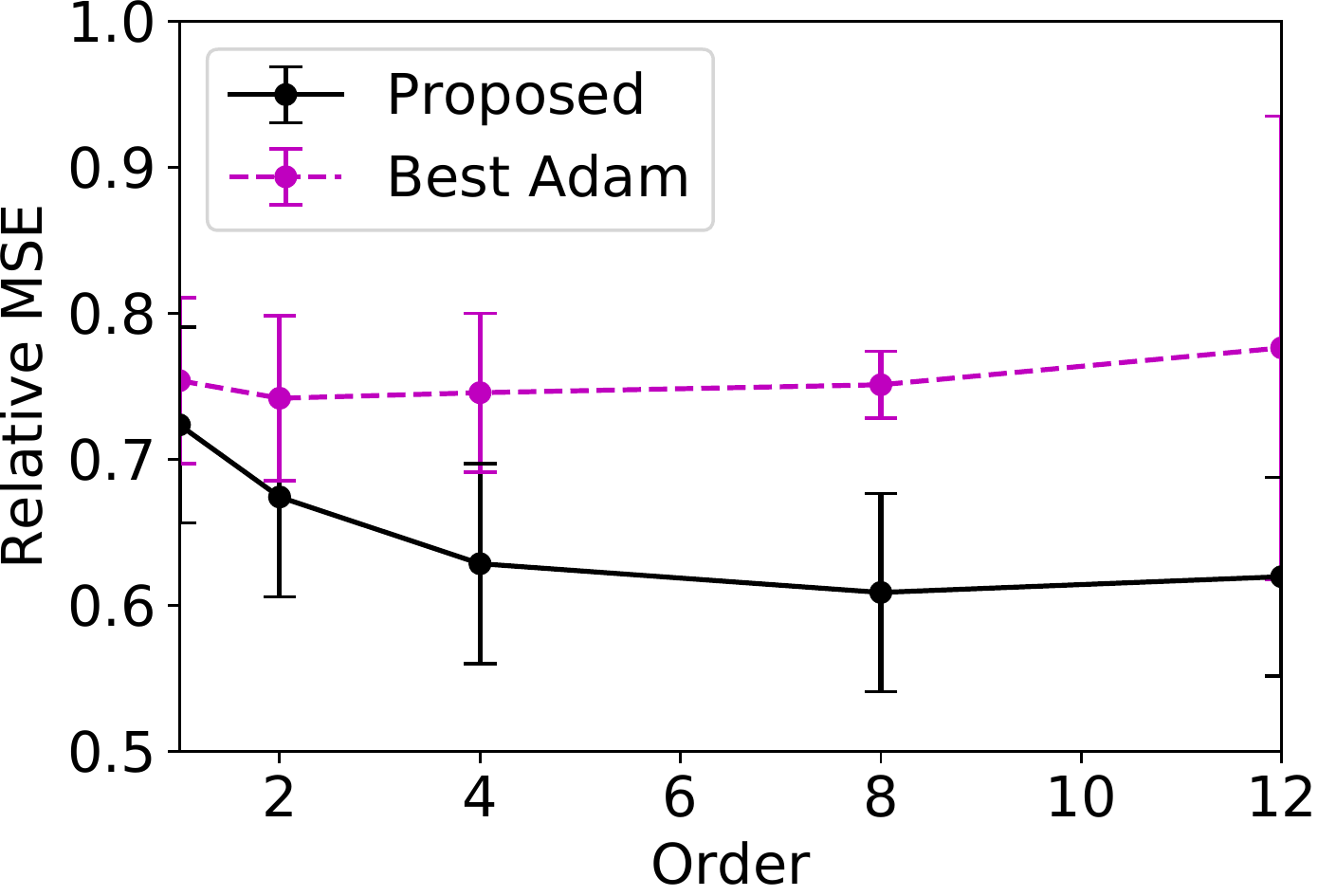}
  \includegraphics[width=0.33\linewidth]{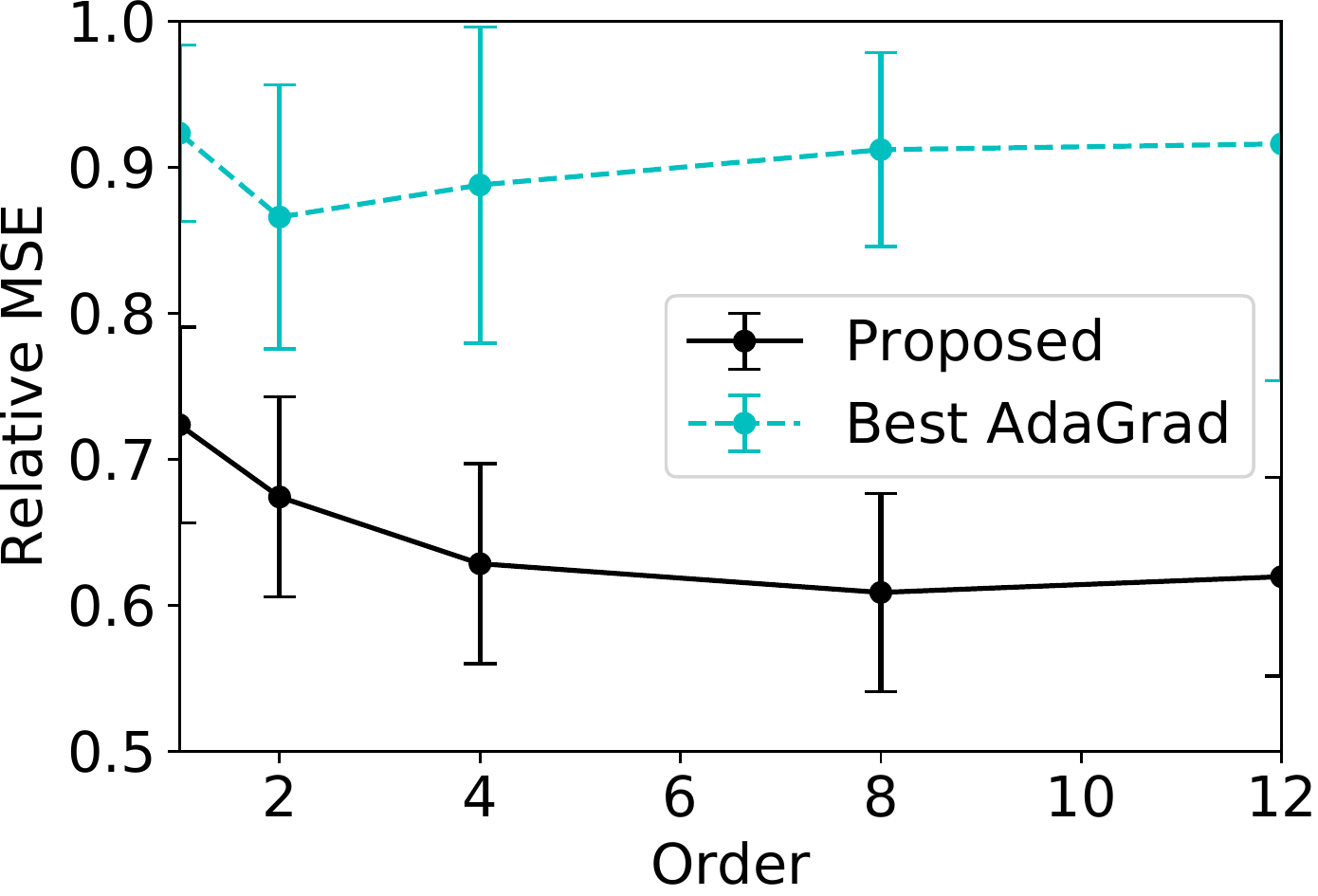}
  \includegraphics[width=0.33\linewidth]{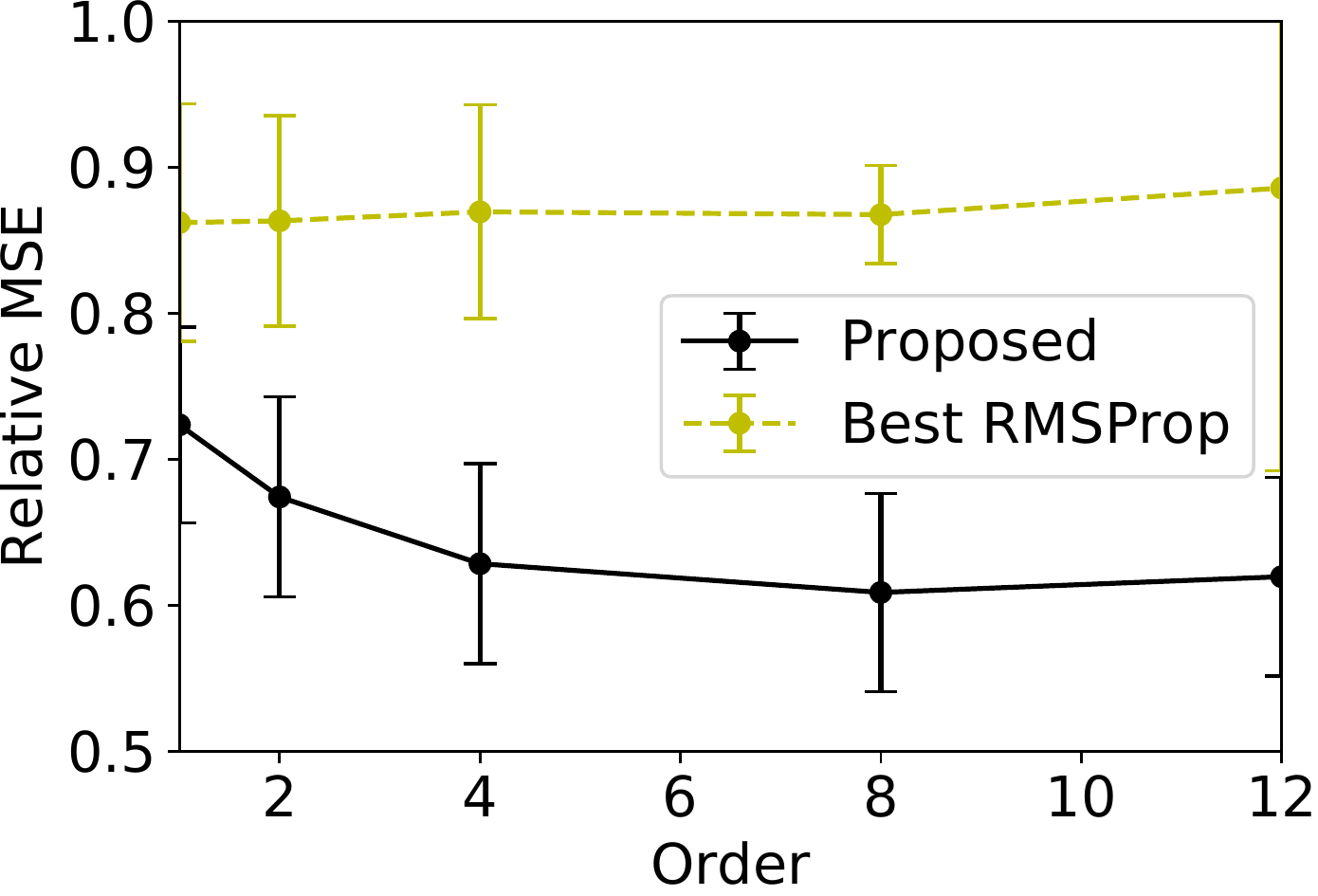}
  \caption{Details of the results on SPX shown in Figure~\ref{fig:all}.
    Each panel shows the relative MSE of Algorithm~\ref{alg:adaptive}
    and a baseline (as indicated in the legend).  The error bars are
    drawn on the basis of the standard deviation of the MSE on each of
    the 10 intervals of equal length.}
  \label{fig:spx}
\end{figure*}

\begin{figure*}[t]
  \begin{centering}
    \underline{\Large Detailed results on Nikkei 225}\\
    \ \\
  \end{centering}
  \includegraphics[width=0.33\linewidth]{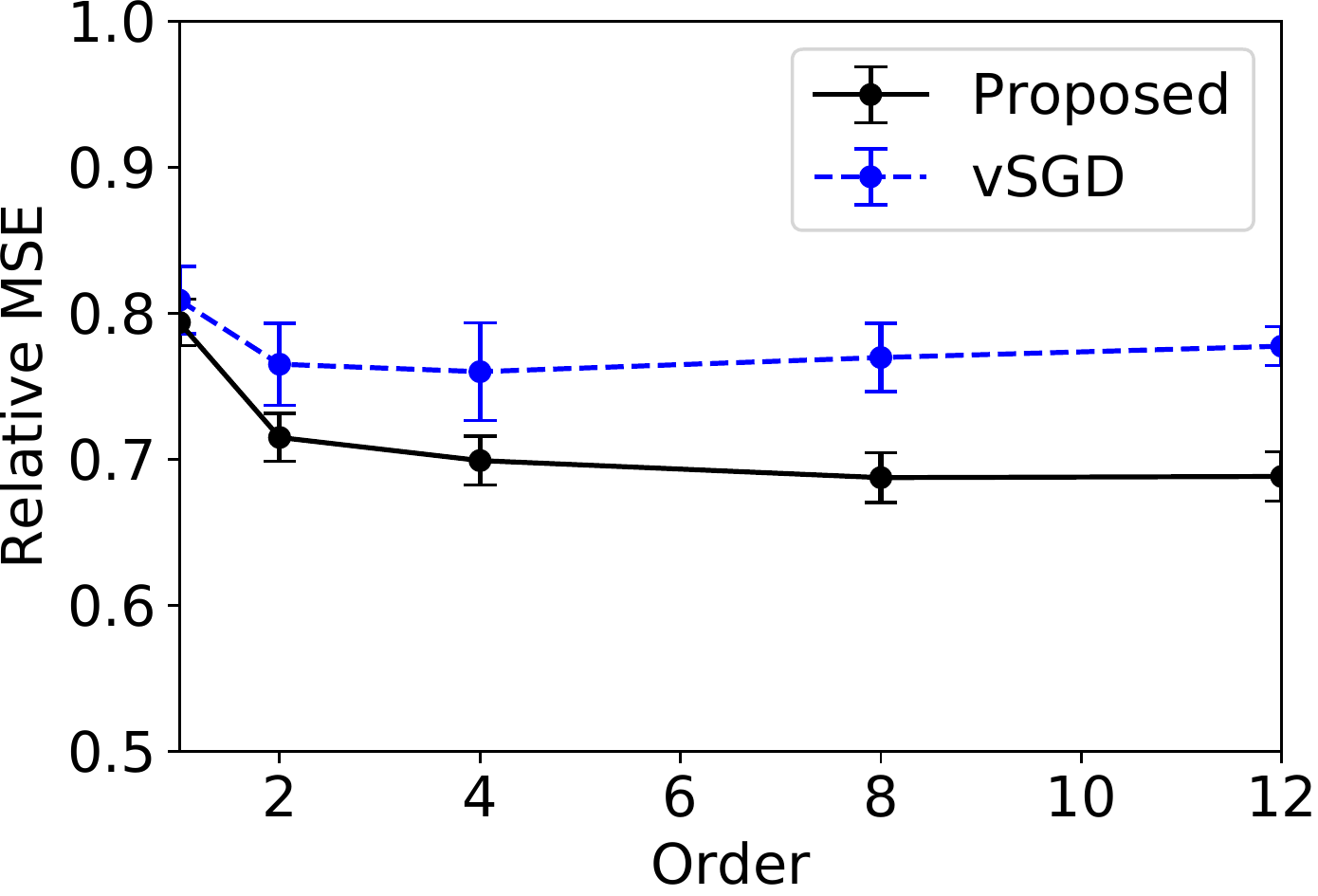}
  \includegraphics[width=0.33\linewidth]{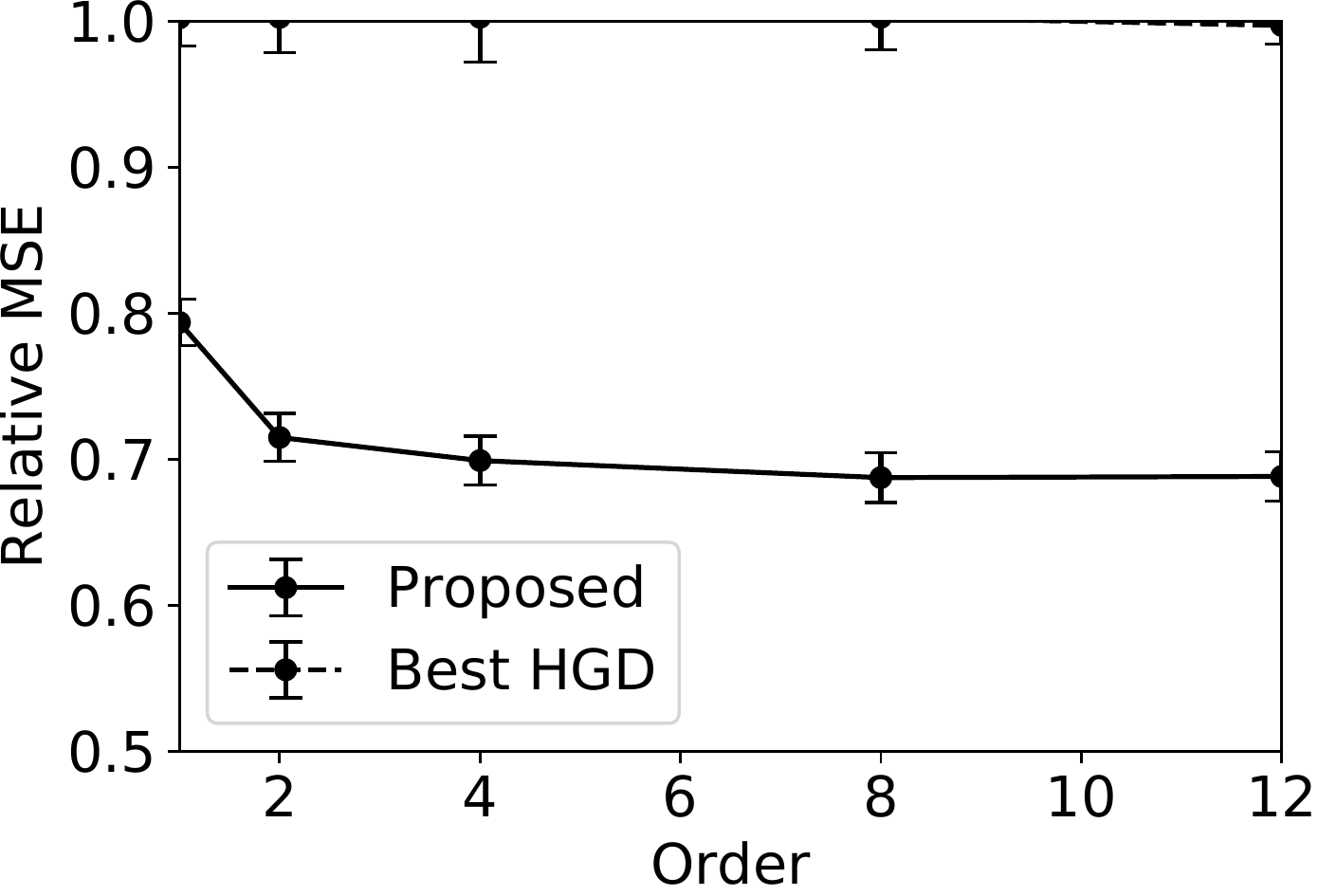}
  \includegraphics[width=0.33\linewidth]{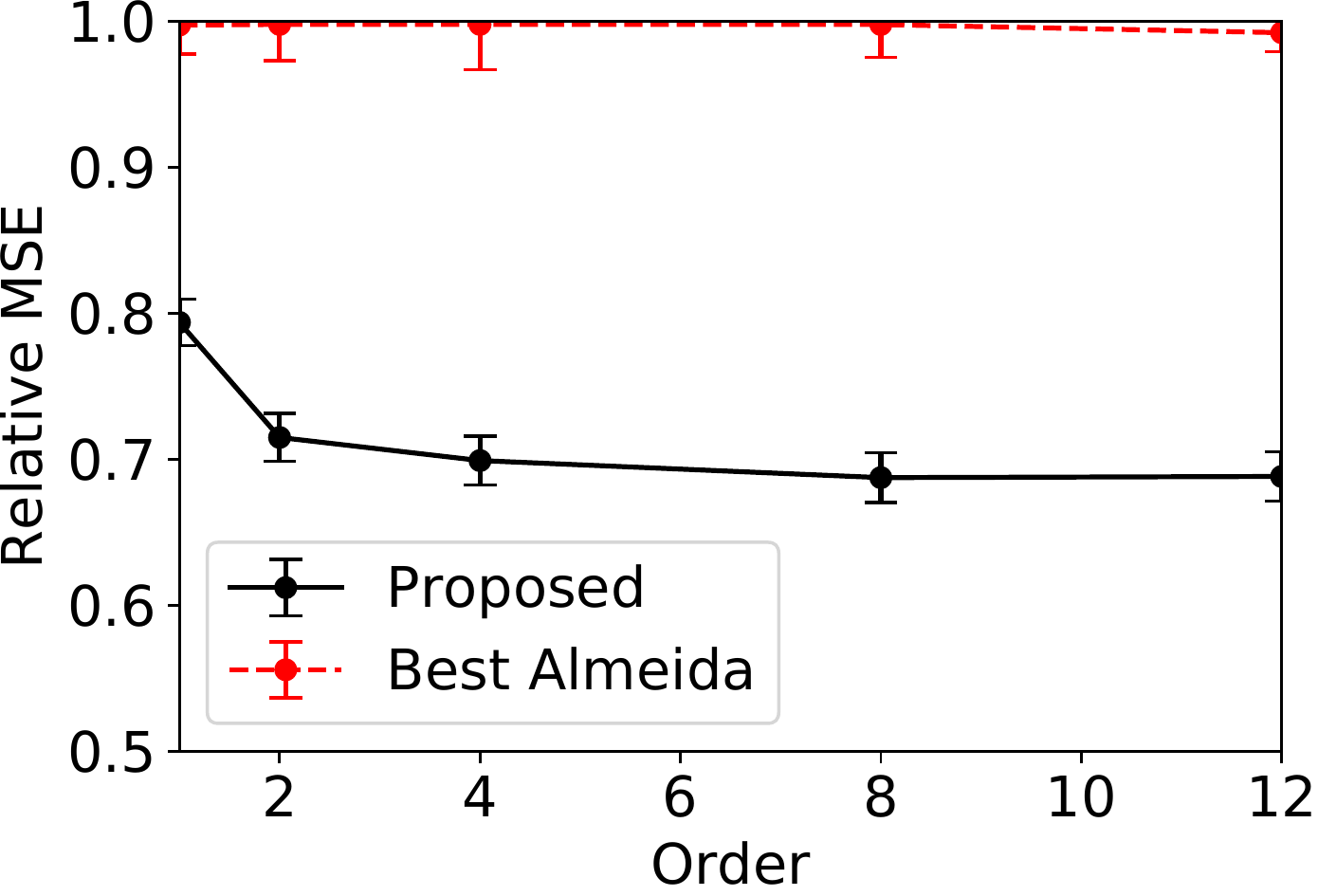}
  \includegraphics[width=0.33\linewidth]{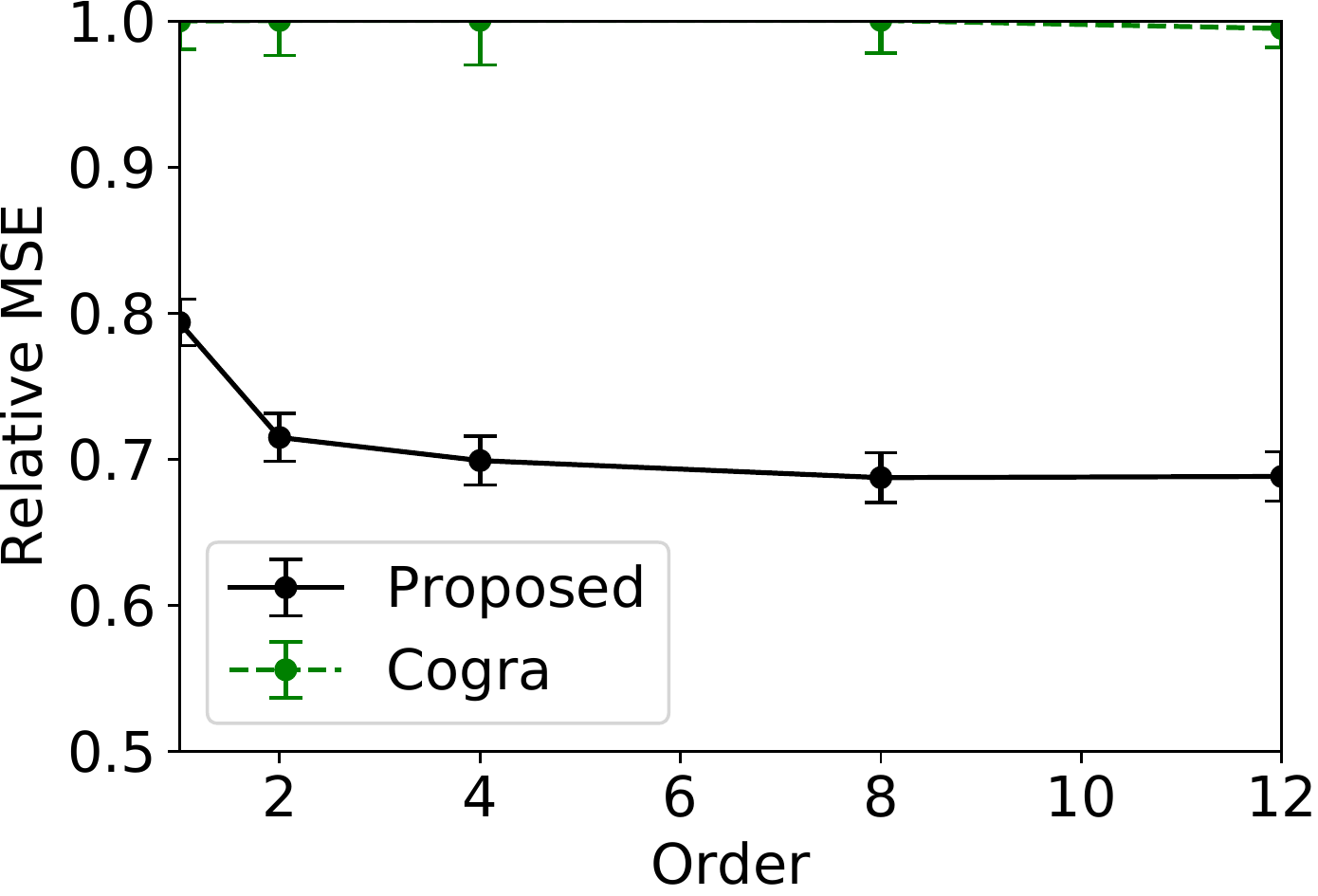}
  \includegraphics[width=0.33\linewidth]{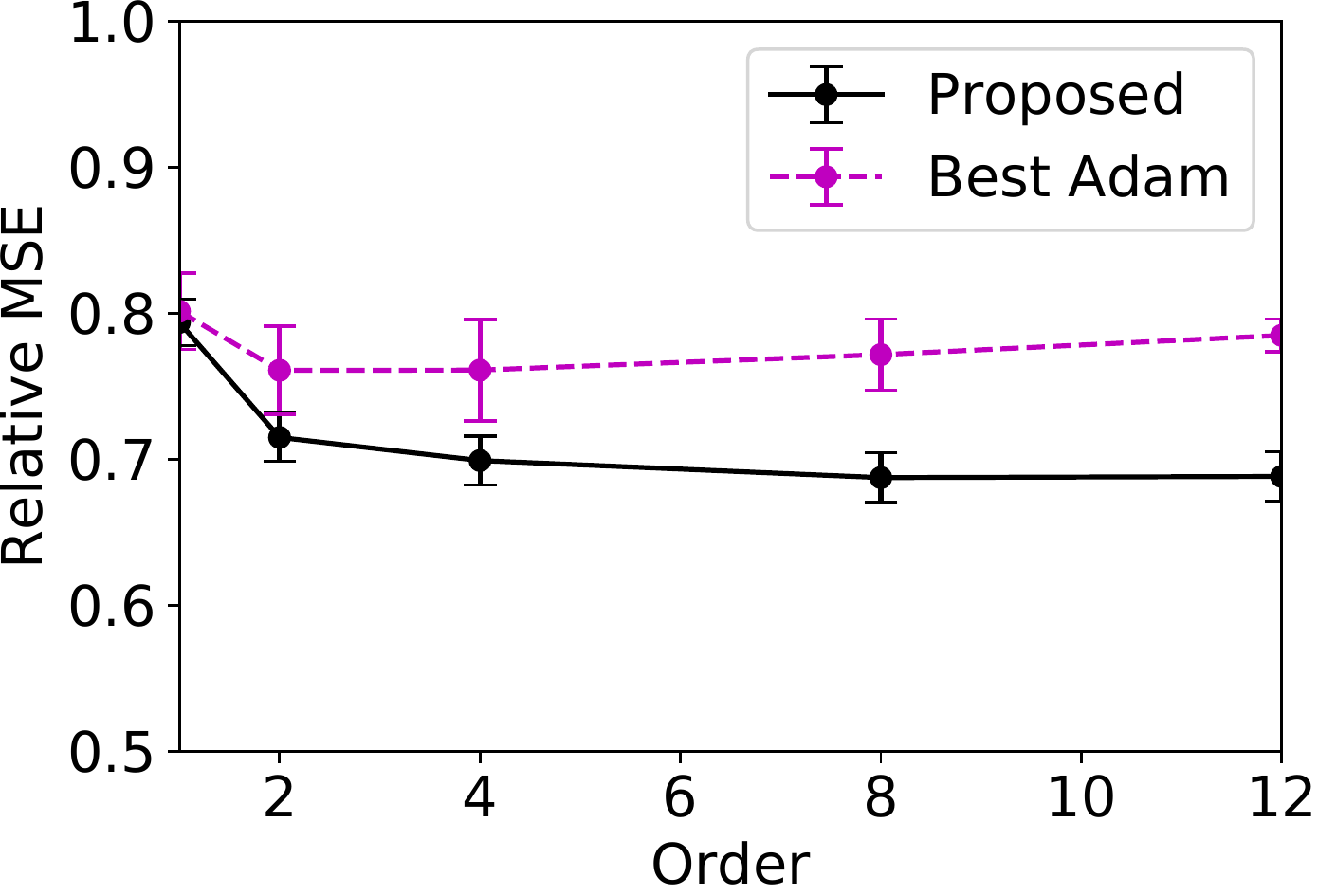}
  \includegraphics[width=0.33\linewidth]{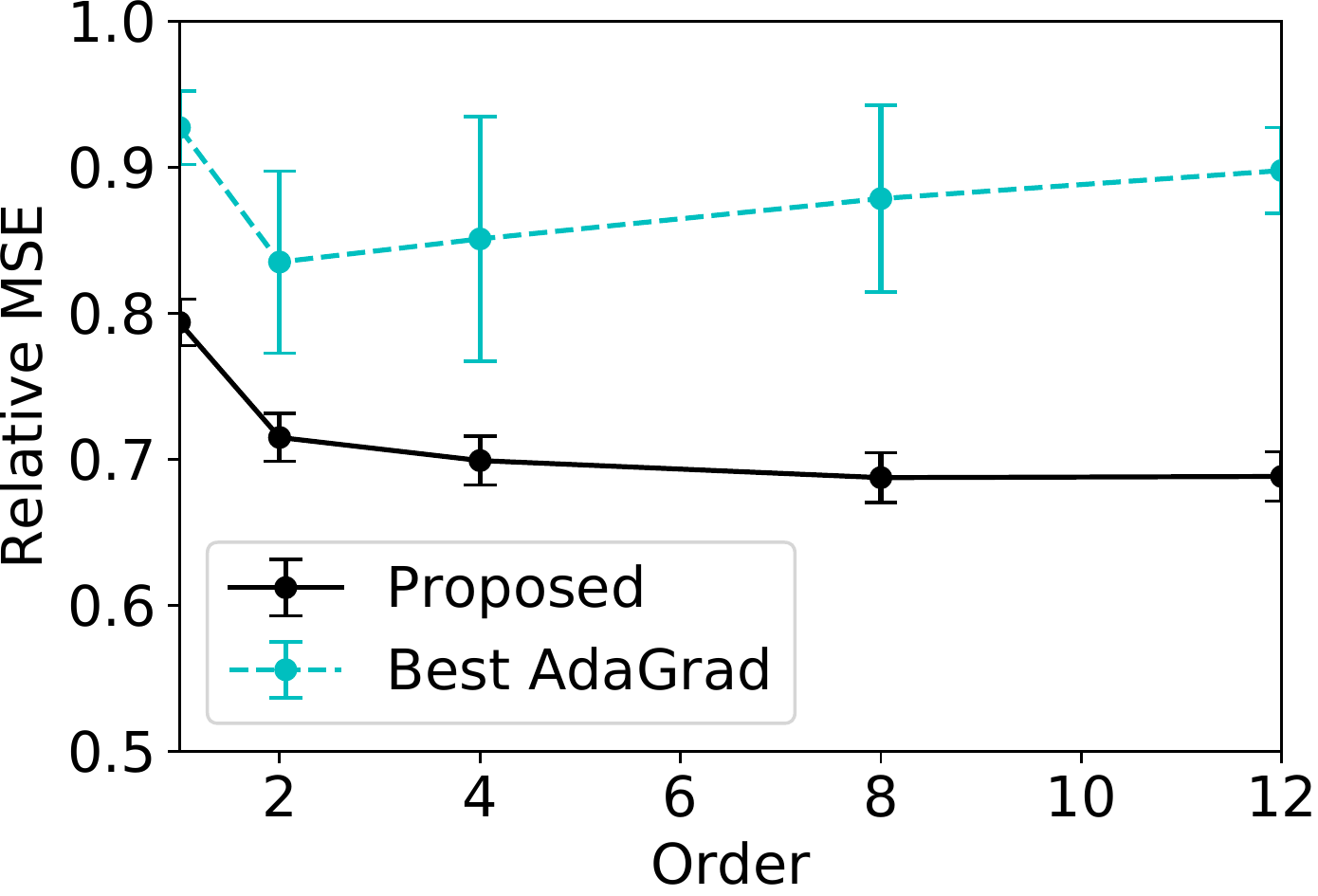}
  \includegraphics[width=0.33\linewidth]{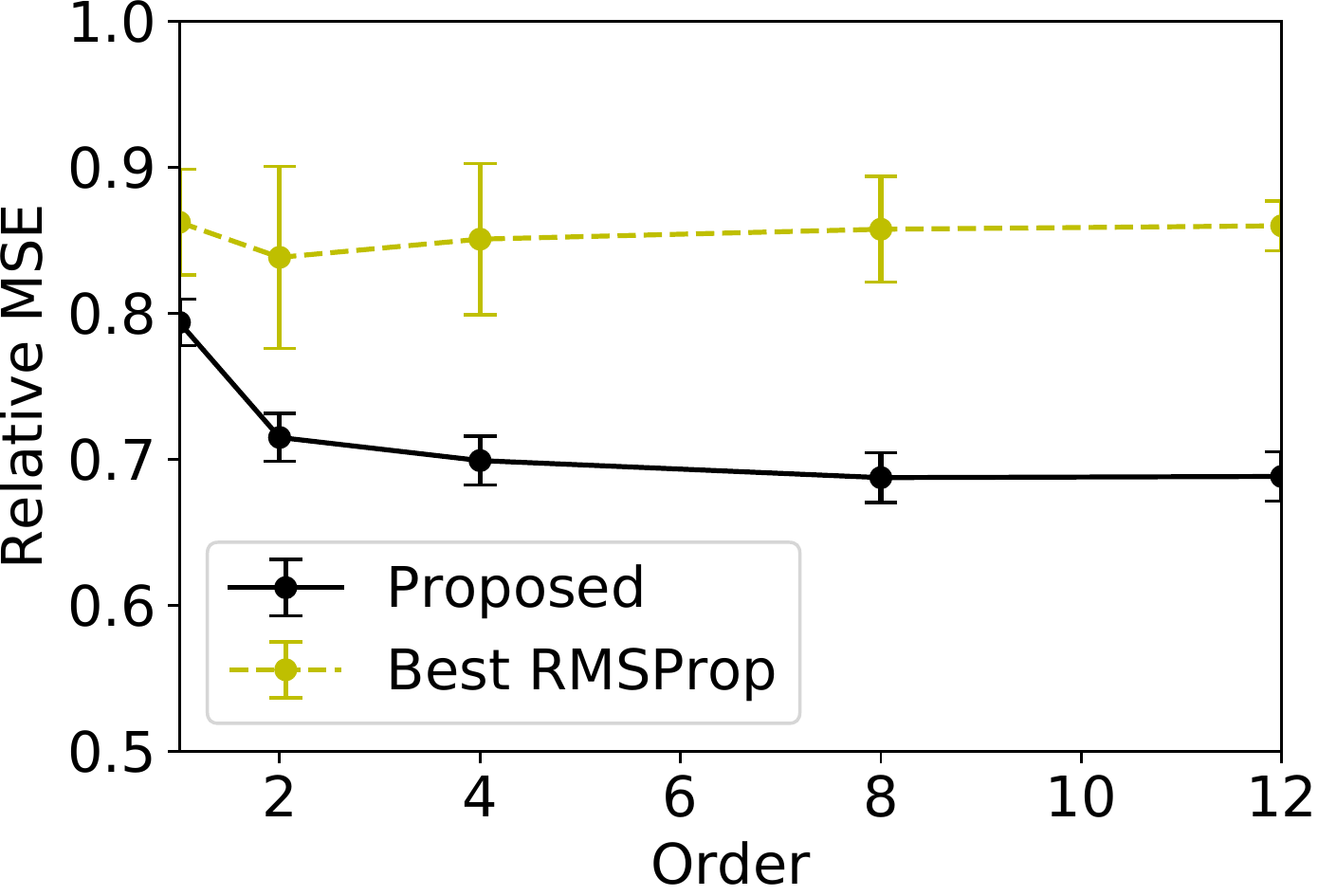}
  \caption{Details of the results on Nikkei 225 shown in Figure~\ref{fig:all}.
    Each panel shows the relative MSE of Algorithm~\ref{alg:adaptive}
    and a baseline (as indicated in the legend).  The error bars are
    drawn on the basis of the standard deviation of the MSE on each of
    the 10 intervals of equal length.}
  \label{fig:nikkei225}
\end{figure*}

\begin{figure*}[t]
  \begin{centering}
    \underline{\Large Detailed results on DAX}\\
    \ \\
  \end{centering}
  \includegraphics[width=0.33\linewidth]{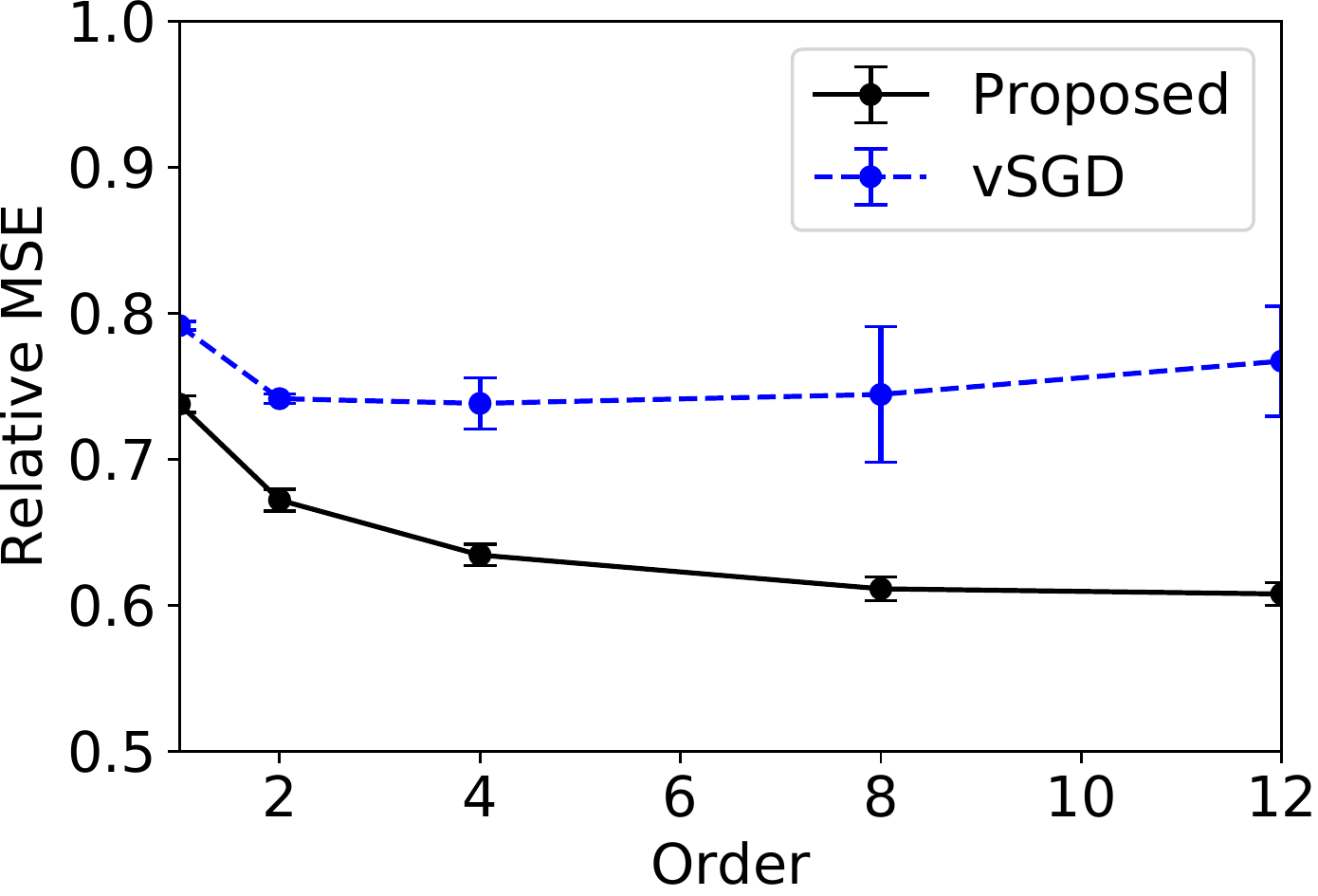}
  \includegraphics[width=0.33\linewidth]{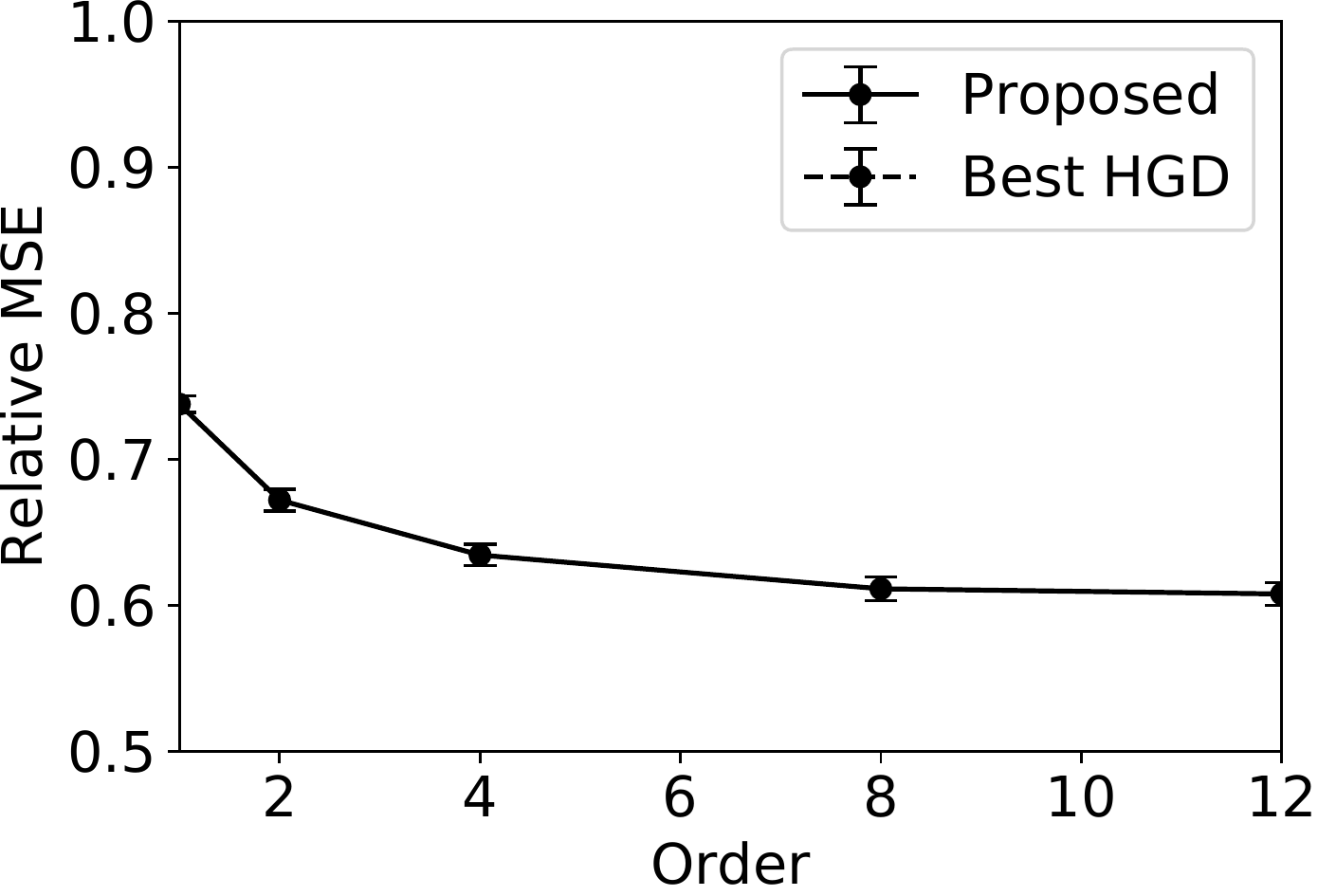}
  \includegraphics[width=0.33\linewidth]{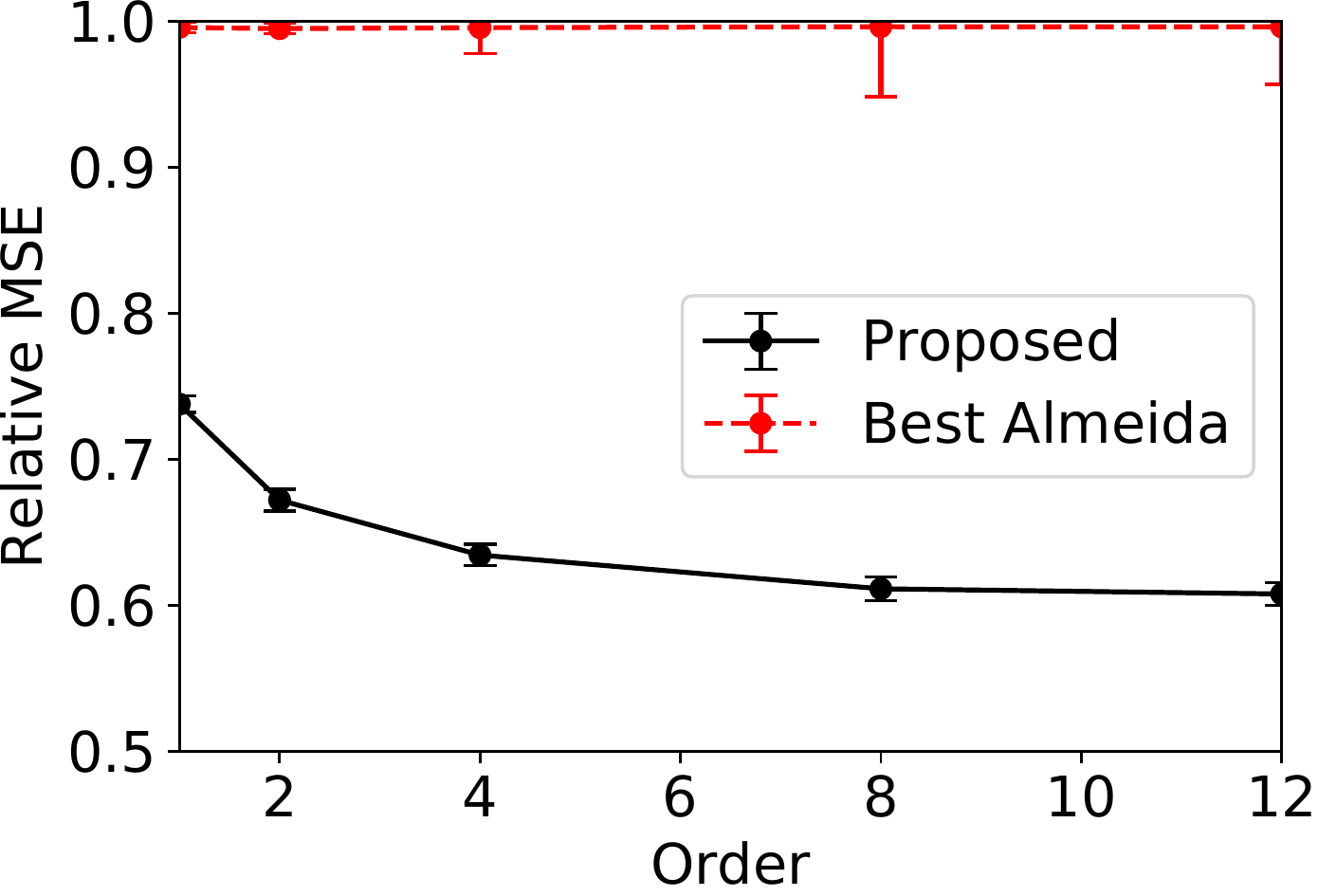}
  \includegraphics[width=0.33\linewidth]{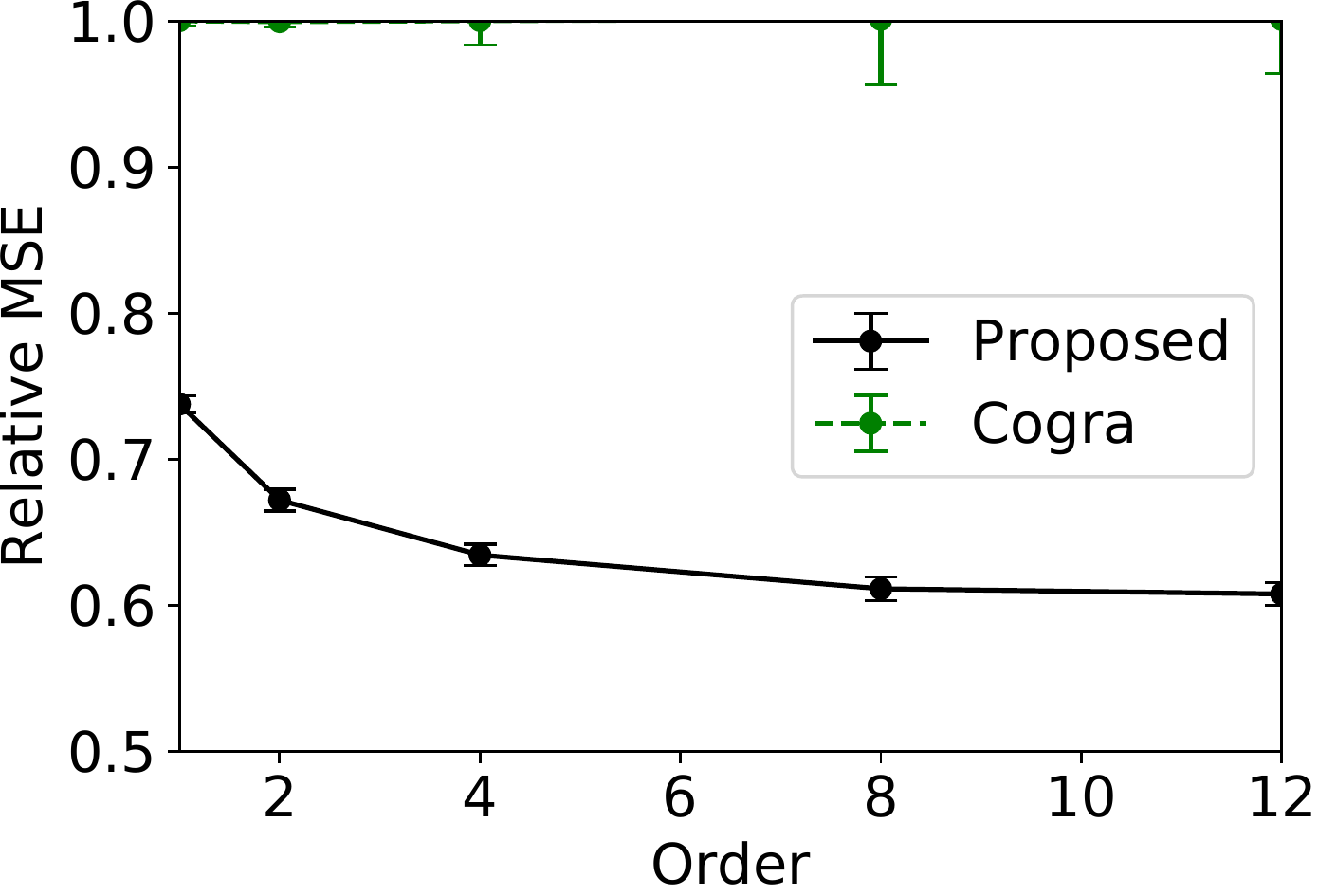}
  \includegraphics[width=0.33\linewidth]{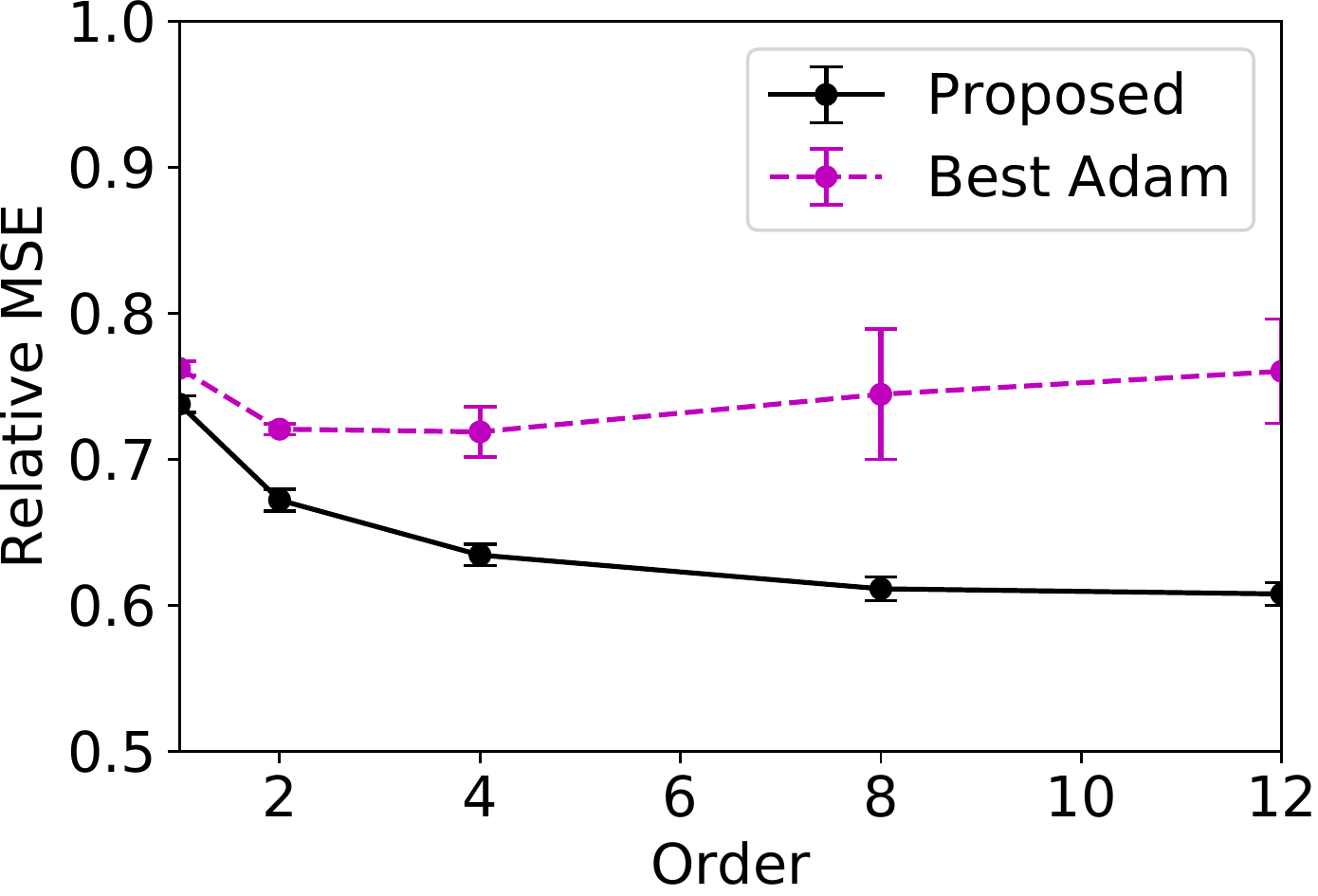}
  \includegraphics[width=0.33\linewidth]{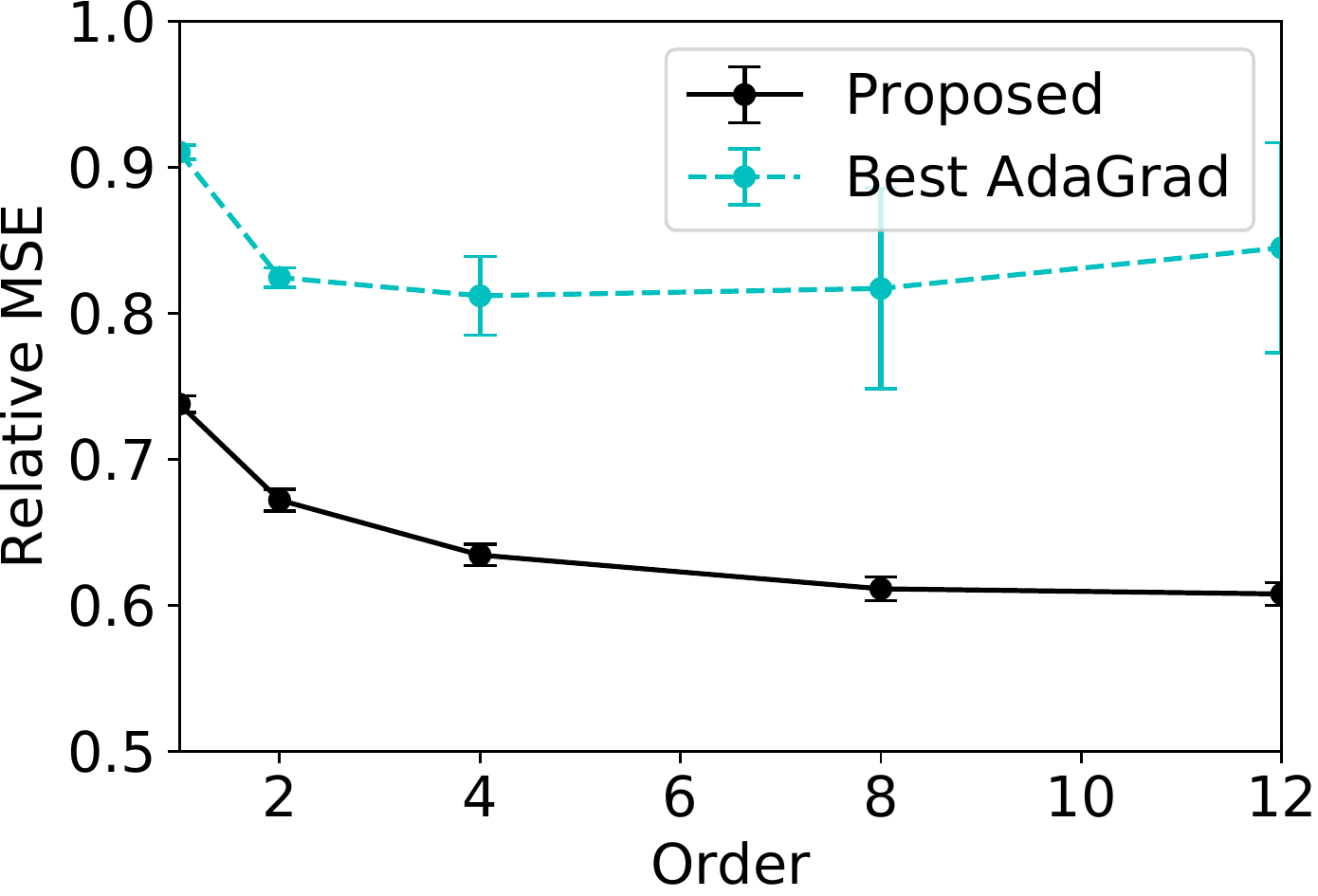}
  \includegraphics[width=0.33\linewidth]{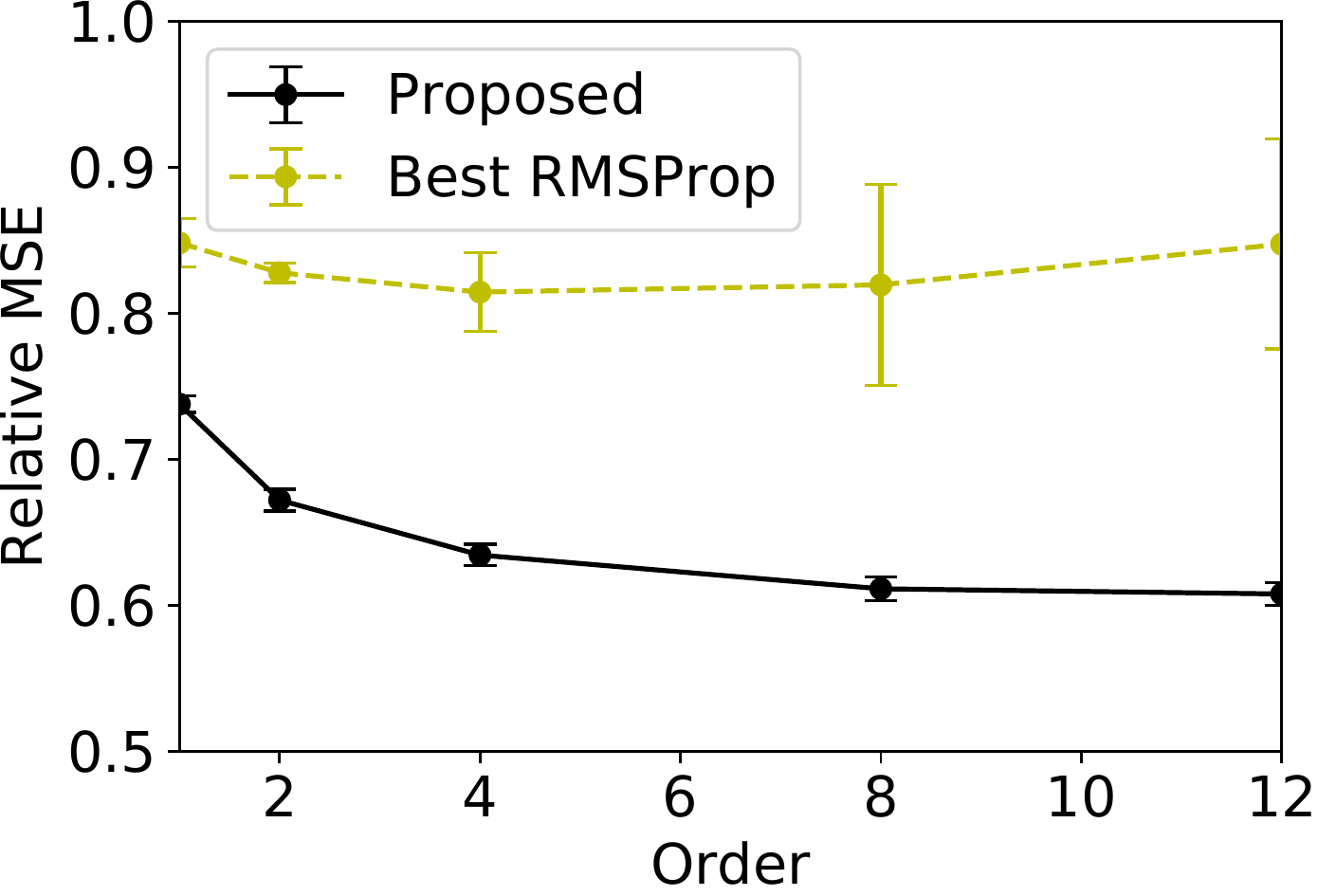}
  \caption{Details of the results on DAX shown in Figure~\ref{fig:all}.
    Each panel shows the relative MSE of Algorithm~\ref{alg:adaptive}
    and a baseline (as indicated in the legend).  The error bars are
    drawn on the basis of the standard deviation of the MSE on each of
    the 10 intervals of equal length.  The relative MSE of
    HGD does not appear in the figure,
    because the relative MSE is above 1.0 for all cases.}
  \label{fig:dax}
\end{figure*}

\begin{figure*}[t]
  \begin{centering}
    \underline{\Large Detailed results on FTSE 100}\\
    \ \\
  \end{centering}
  \includegraphics[width=0.33\linewidth]{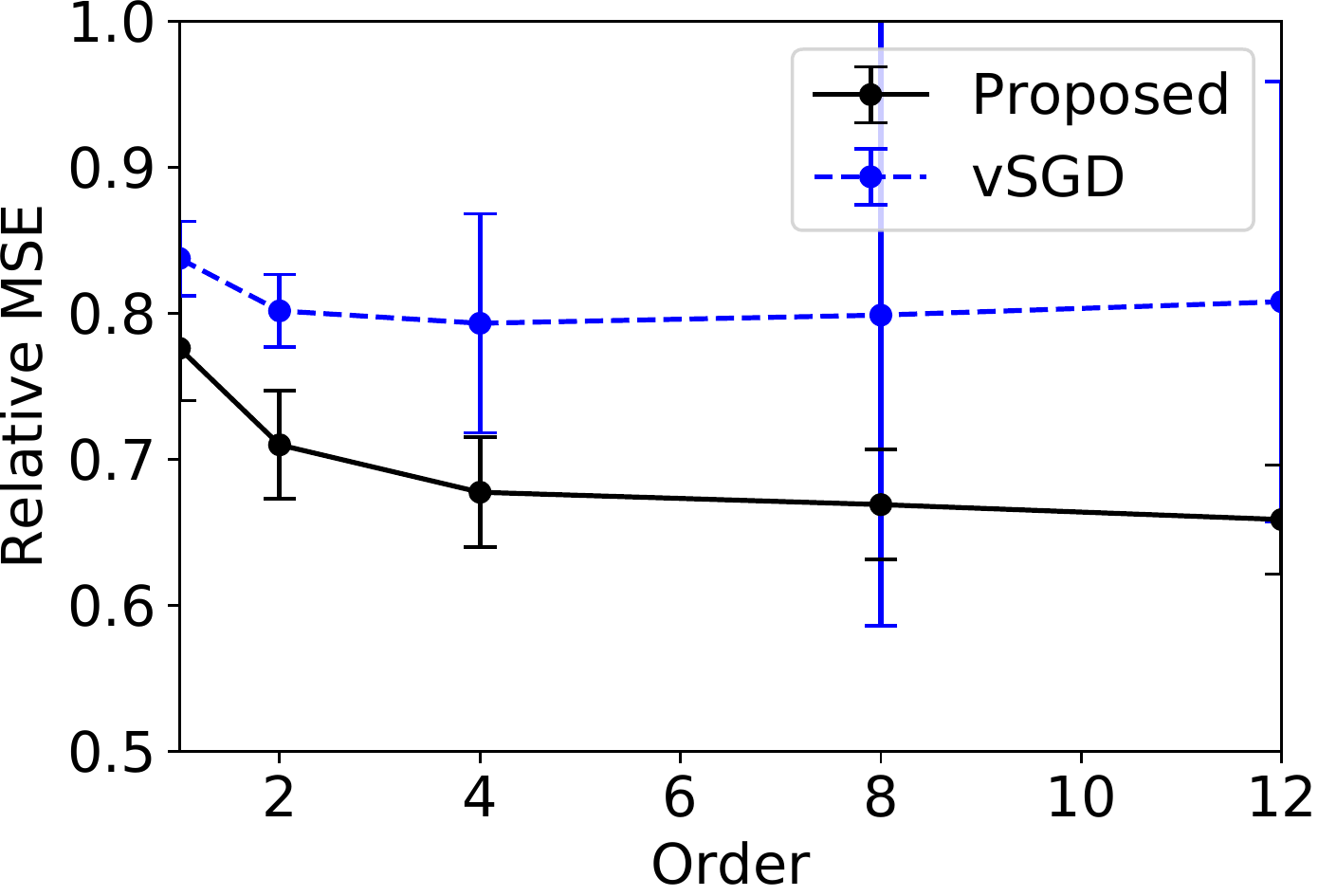}
  \includegraphics[width=0.33\linewidth]{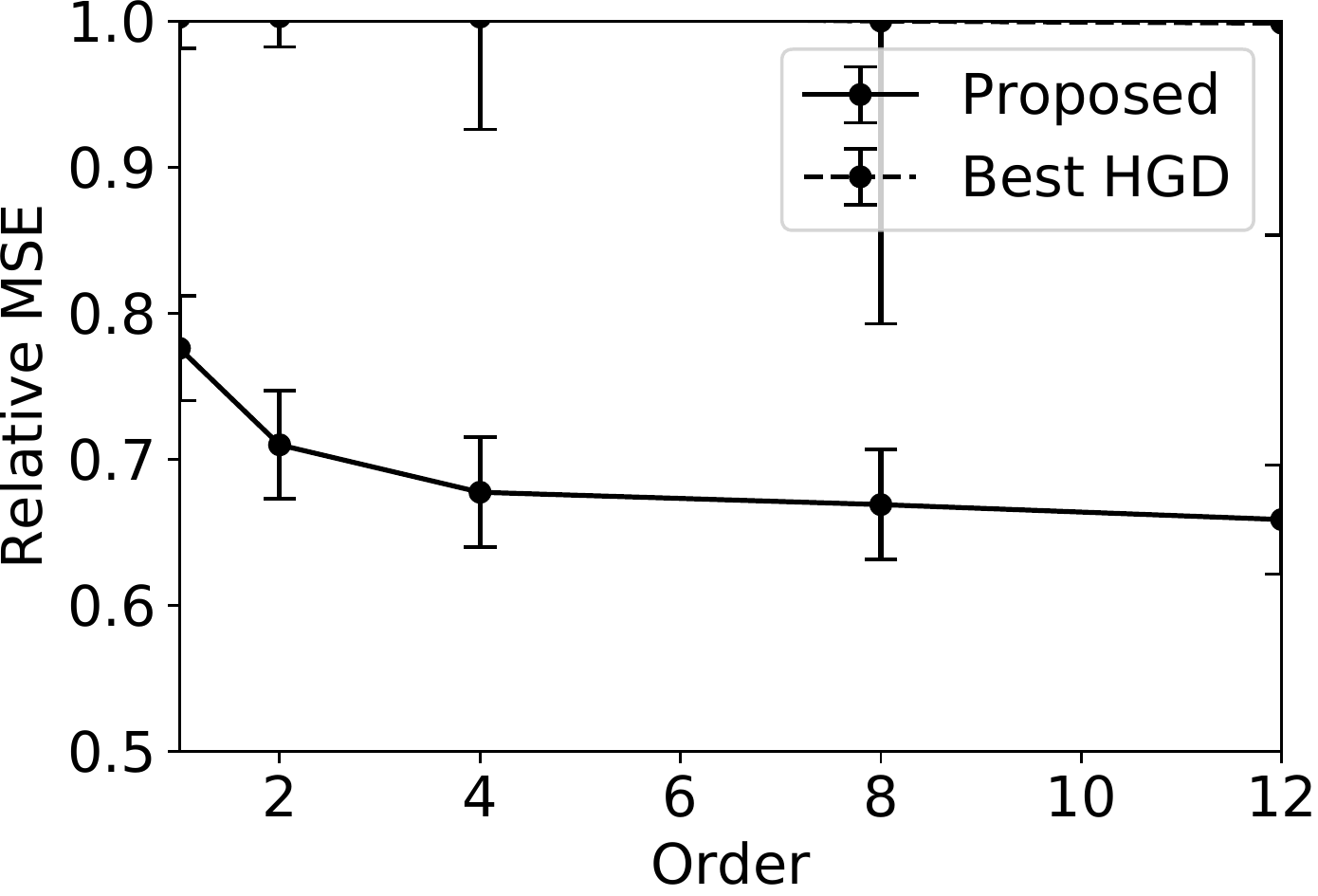}
  \includegraphics[width=0.33\linewidth]{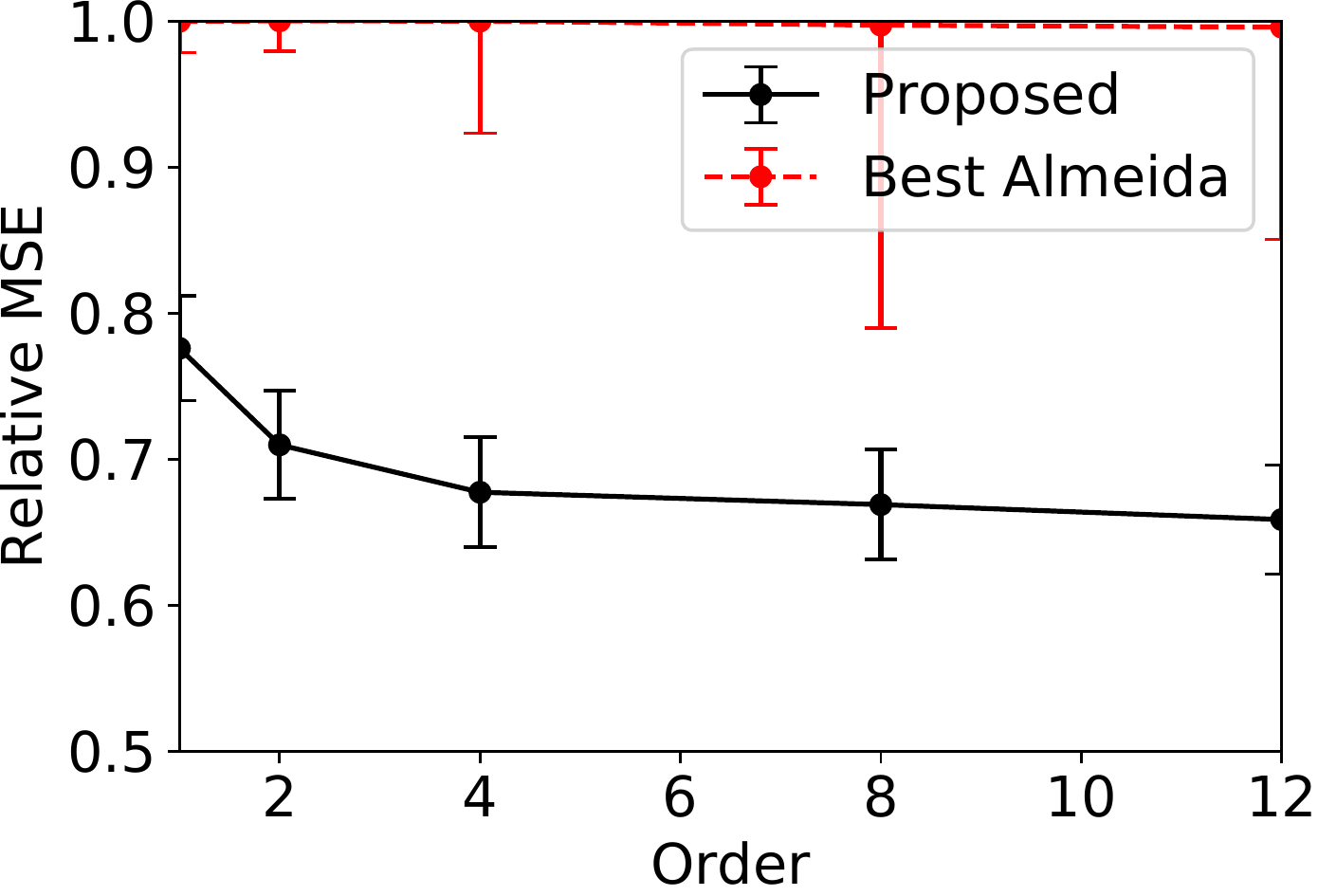}
  \includegraphics[width=0.33\linewidth]{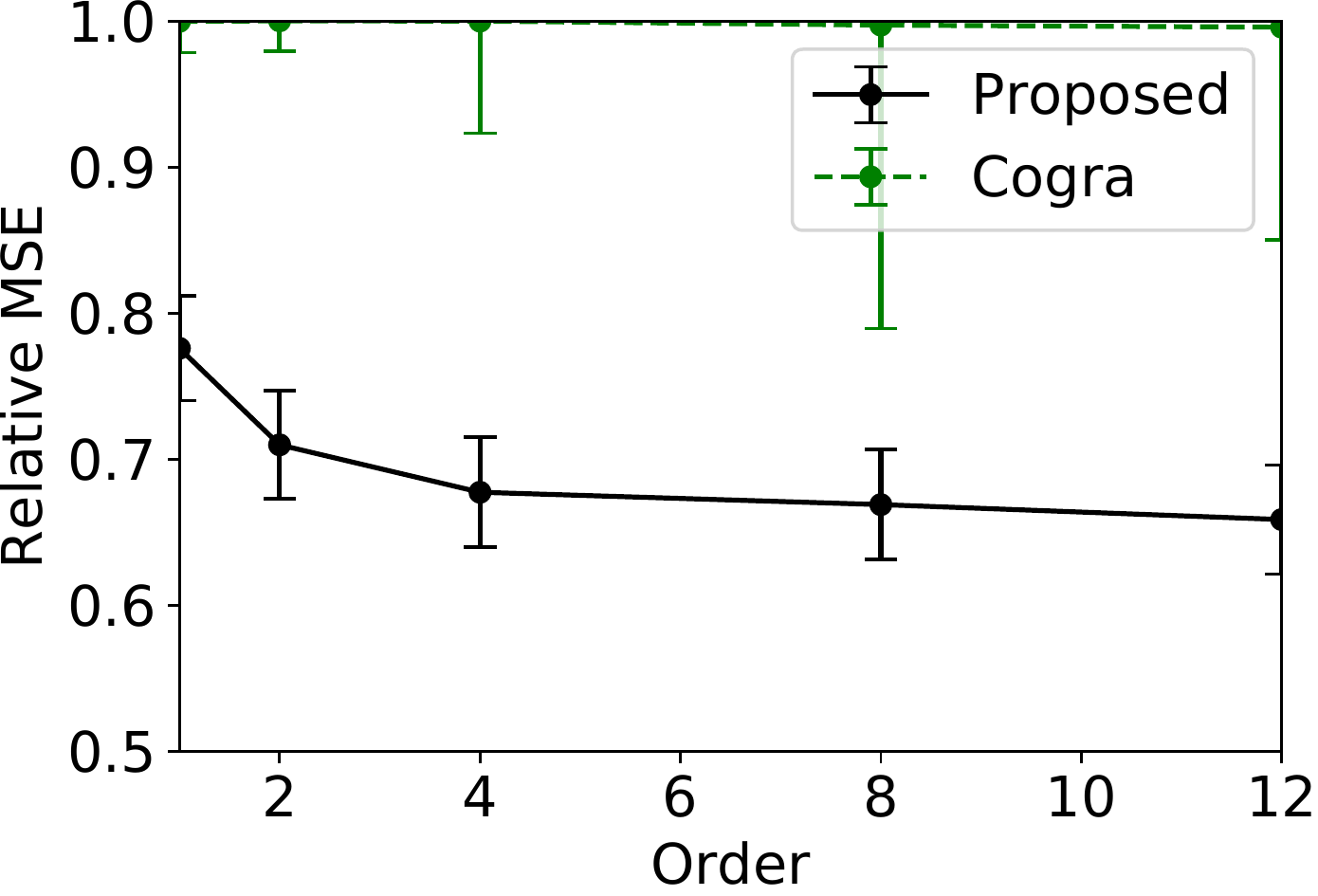}
  \includegraphics[width=0.33\linewidth]{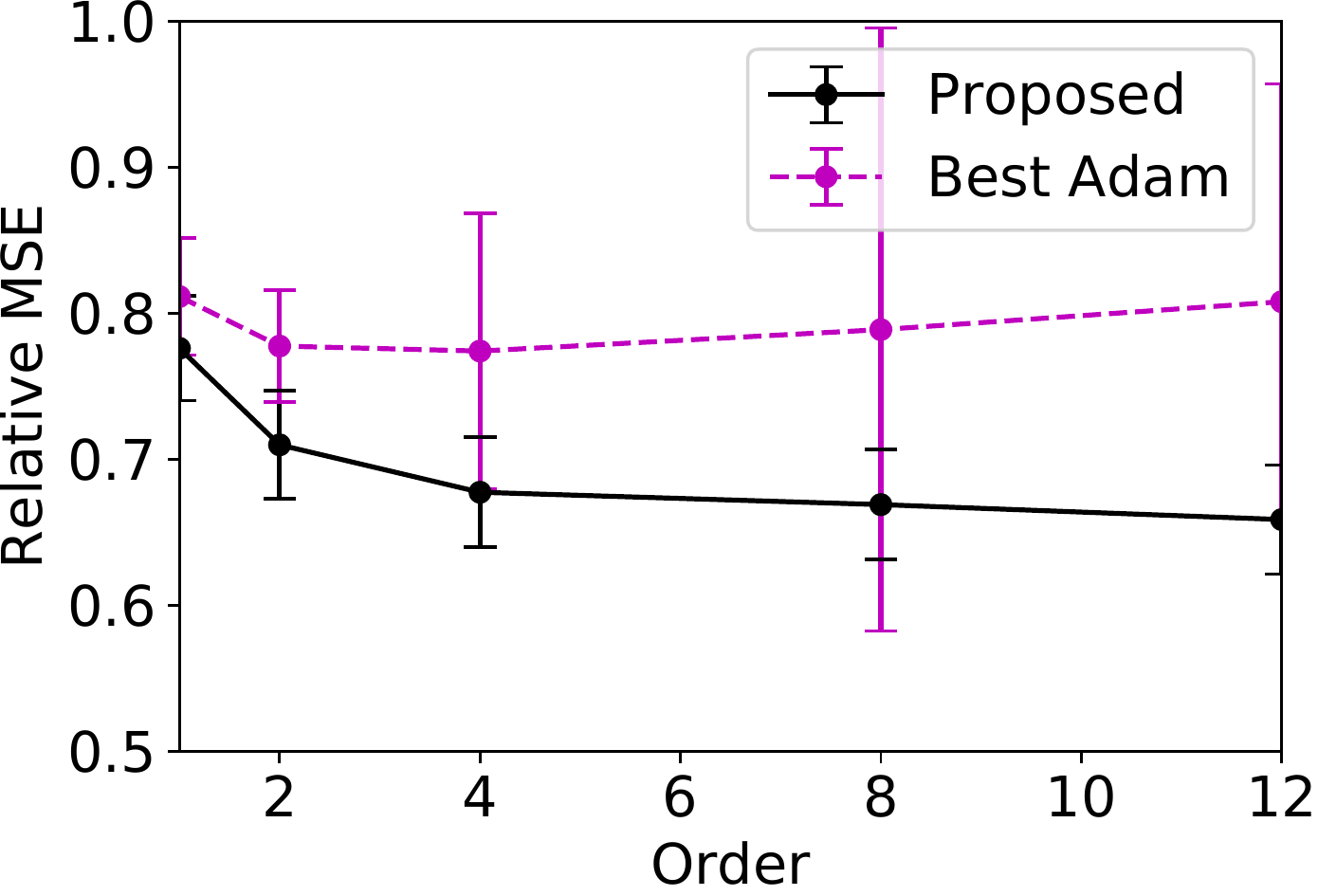}
  \includegraphics[width=0.33\linewidth]{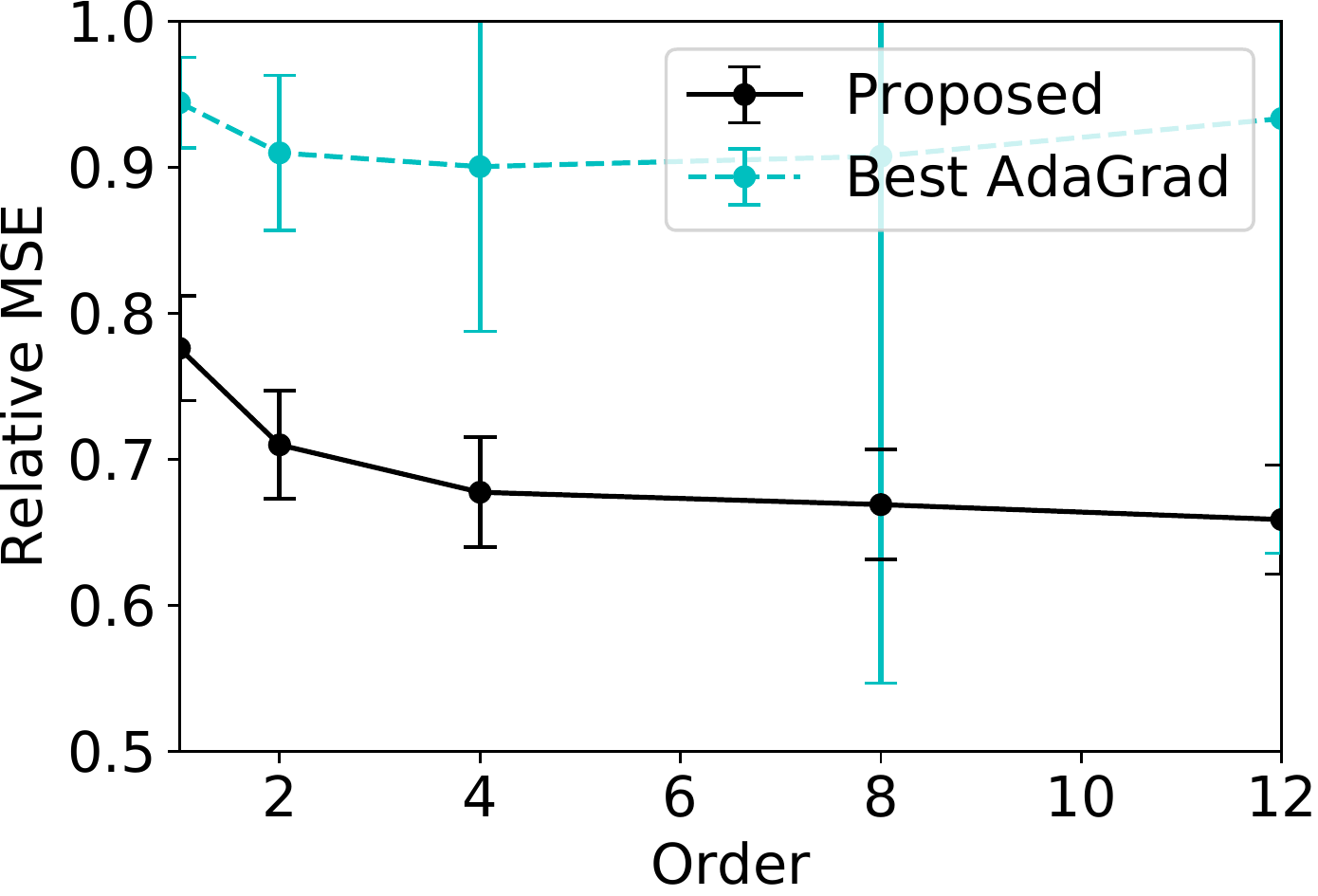}
  \includegraphics[width=0.33\linewidth]{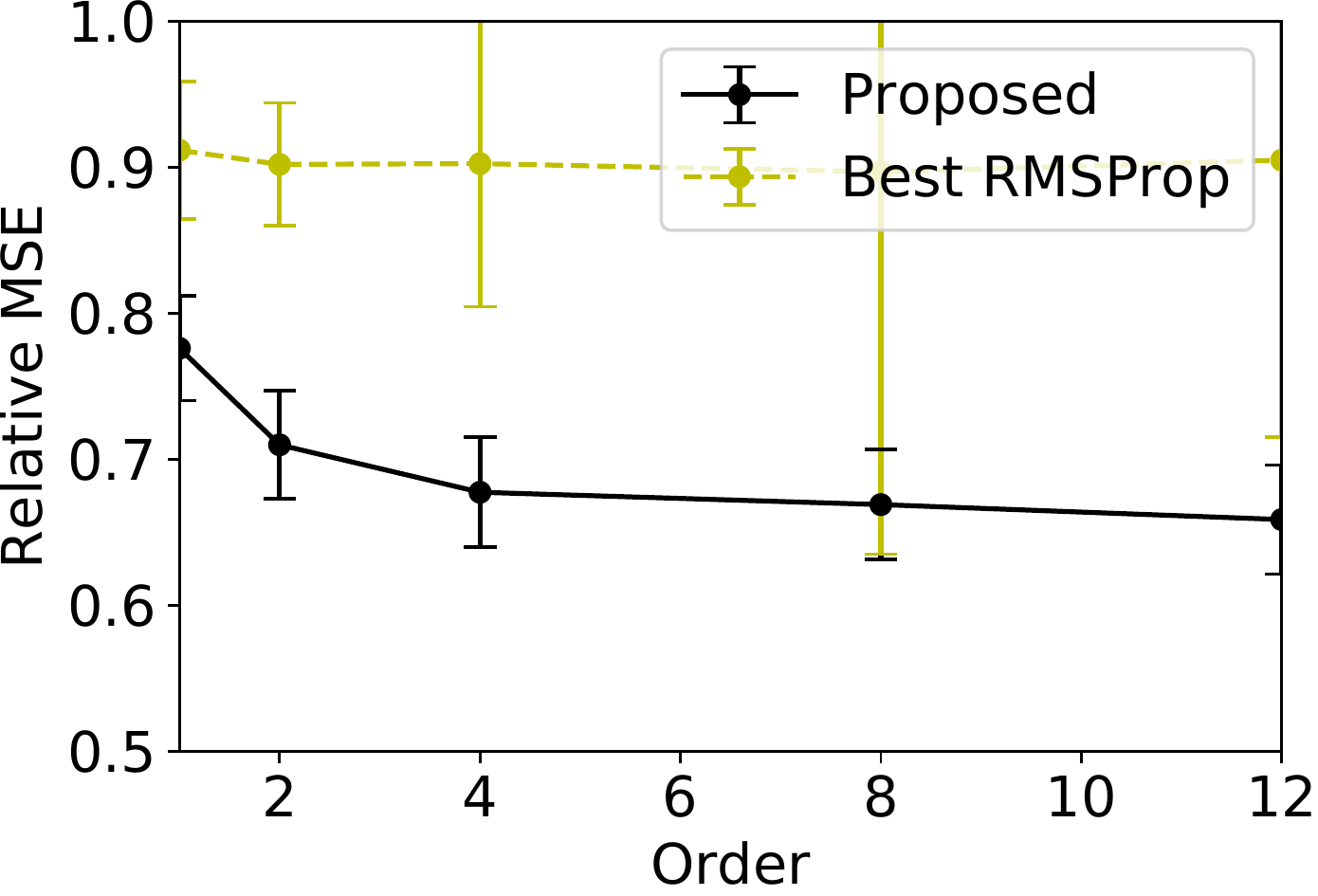}
  \caption{Details of the results on FTSE 100 shown in Figure~\ref{fig:all}.
    Each panel shows the relative MSE of Algorithm~\ref{alg:adaptive}
    and a baseline (as indicated in the legend).  The error bars are
    drawn on the basis of the standard deviation of the MSE on each of
    the 10 intervals of equal length.}
  \label{fig:ftse100}
\end{figure*}

\begin{figure*}[t]
  \begin{centering}
    \underline{\Large Detailed results on SSEC}\\
    \ \\
  \end{centering}
  \includegraphics[width=0.33\linewidth]{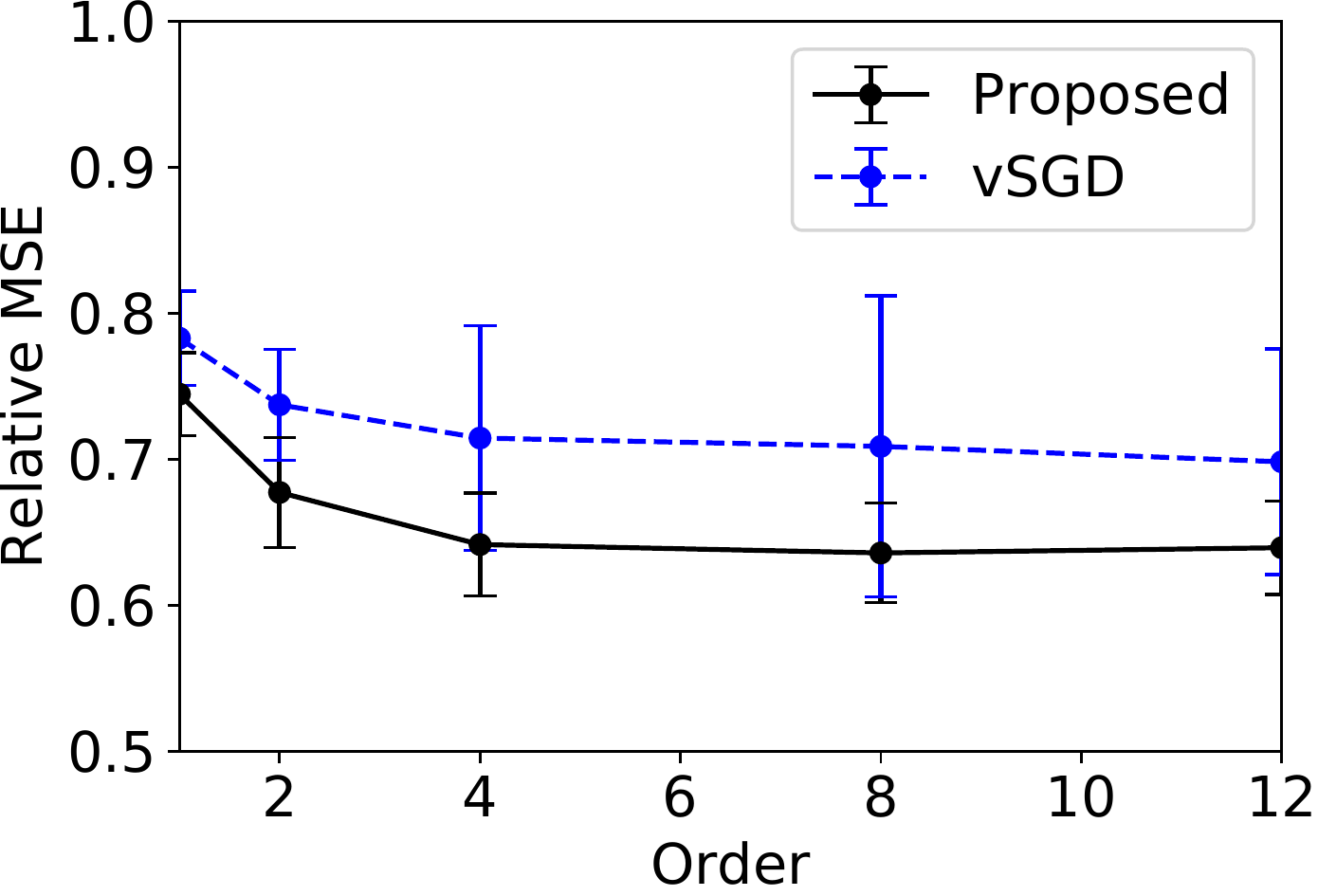}
  \includegraphics[width=0.33\linewidth]{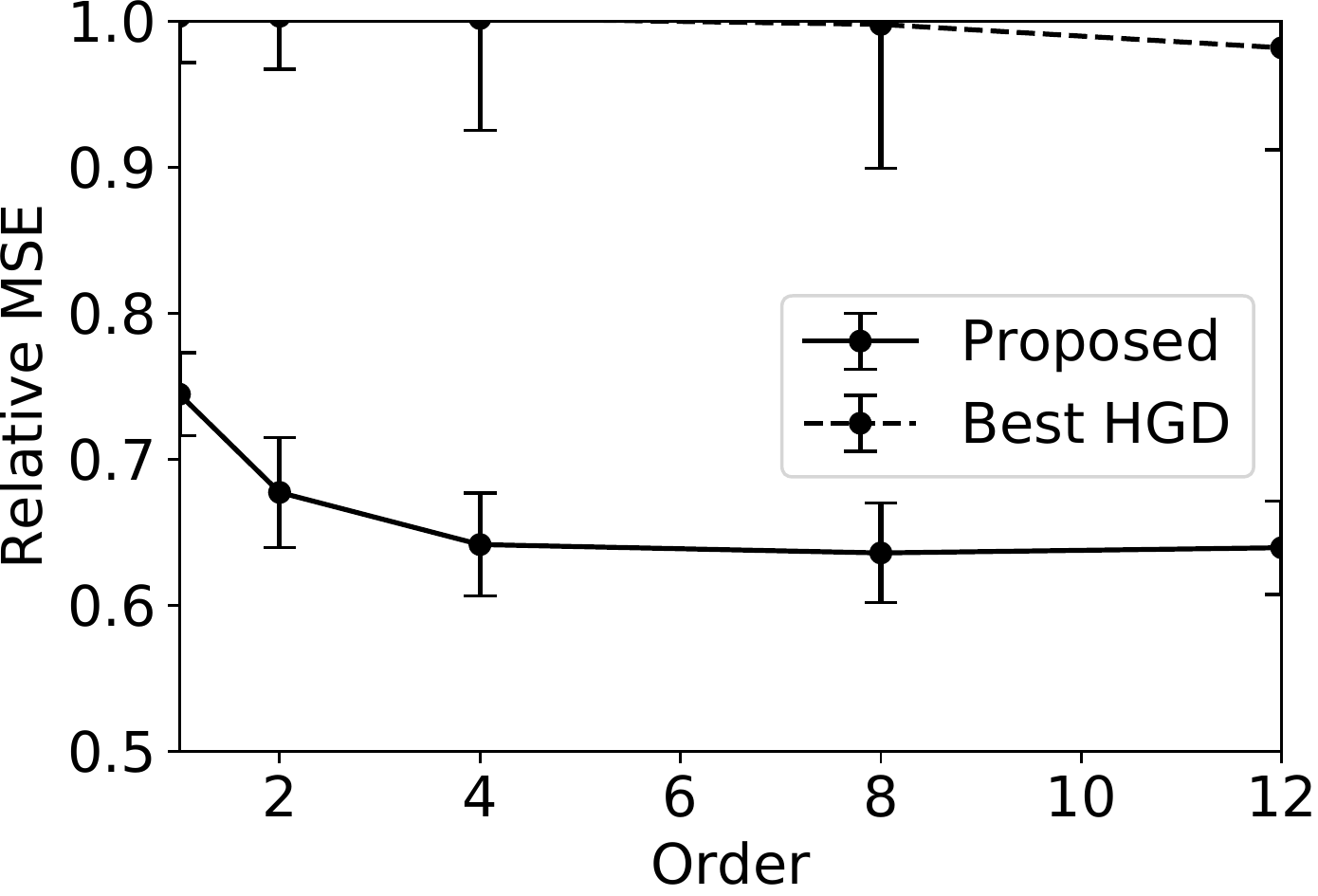}
  \includegraphics[width=0.33\linewidth]{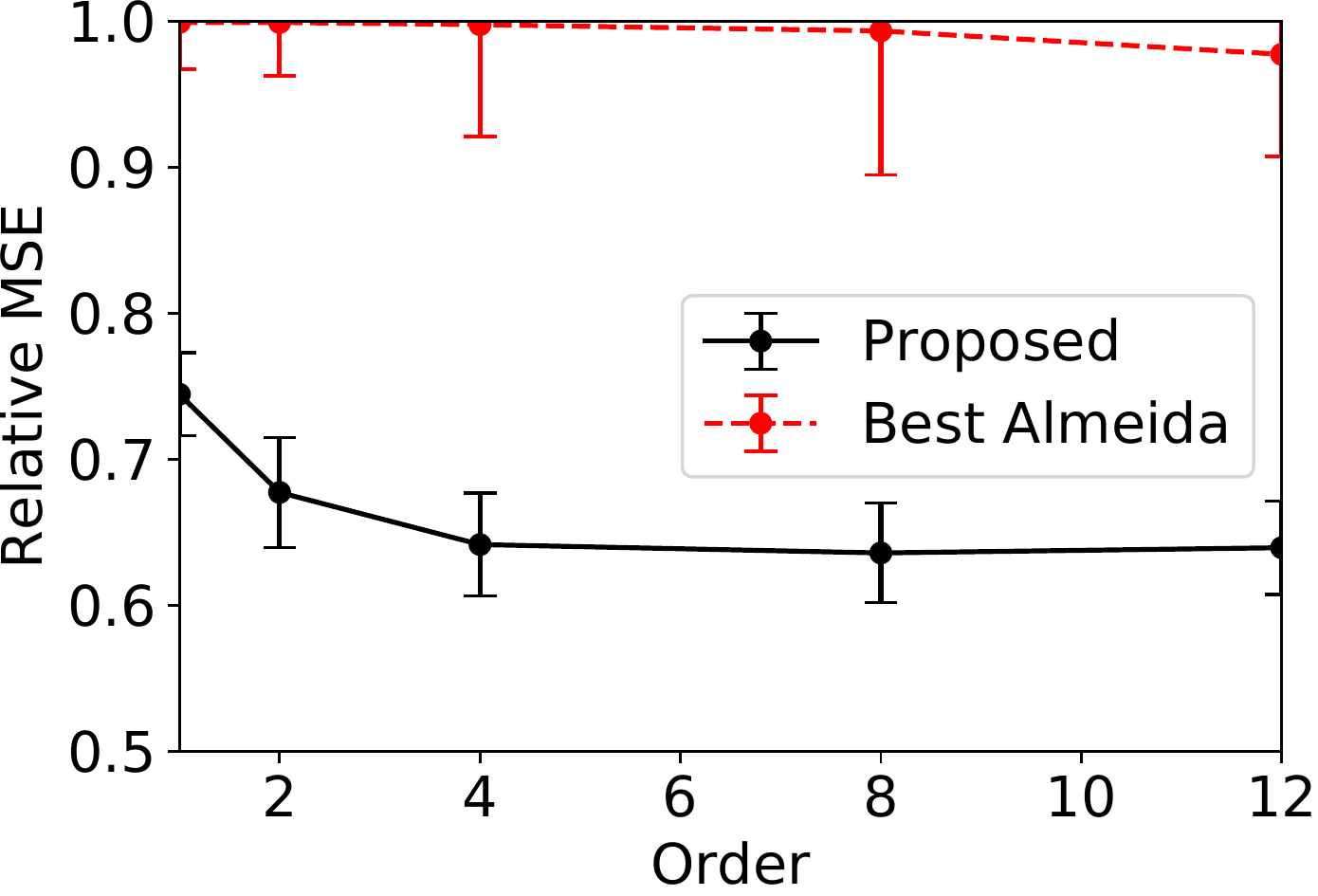}
  \includegraphics[width=0.33\linewidth]{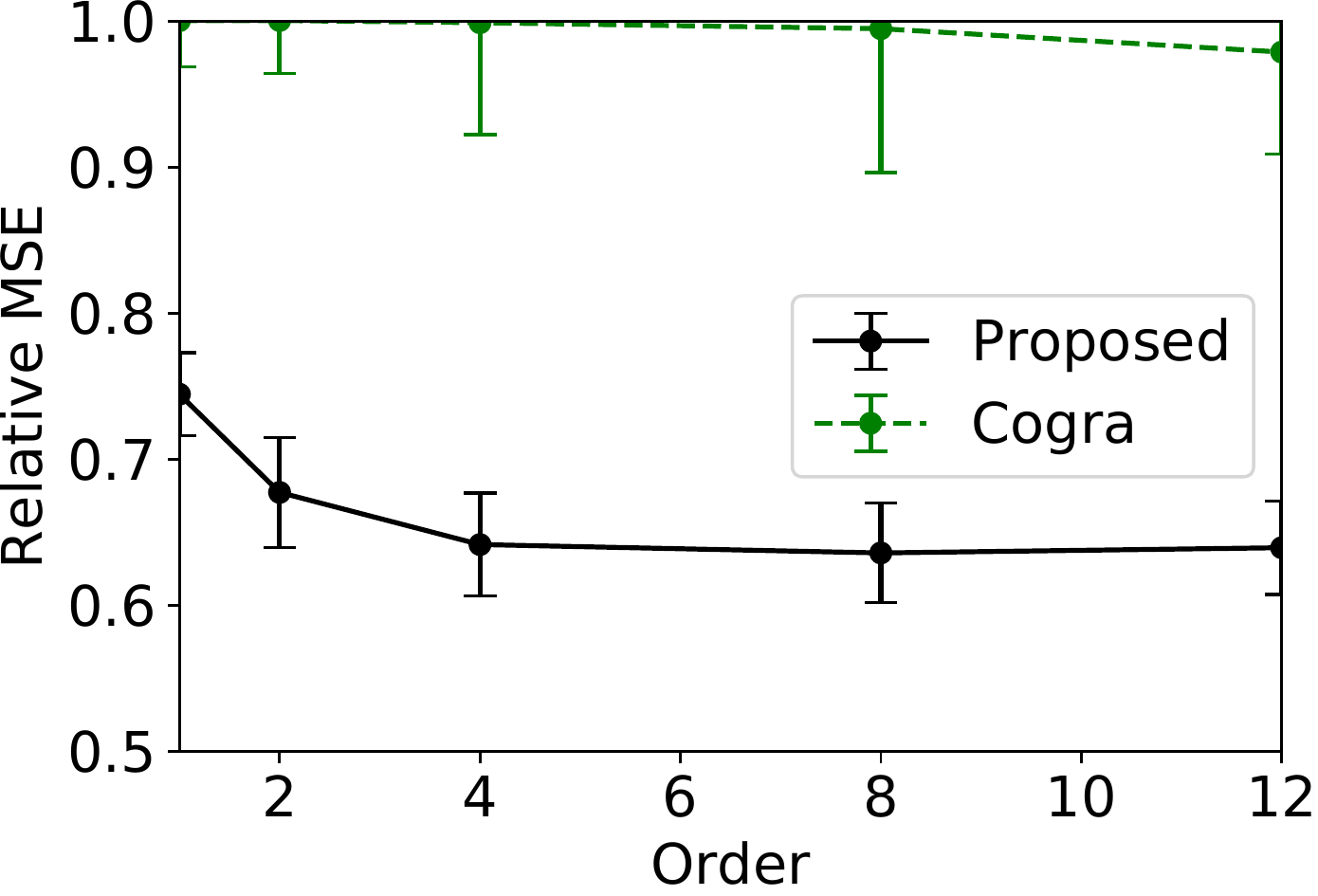}
  \includegraphics[width=0.33\linewidth]{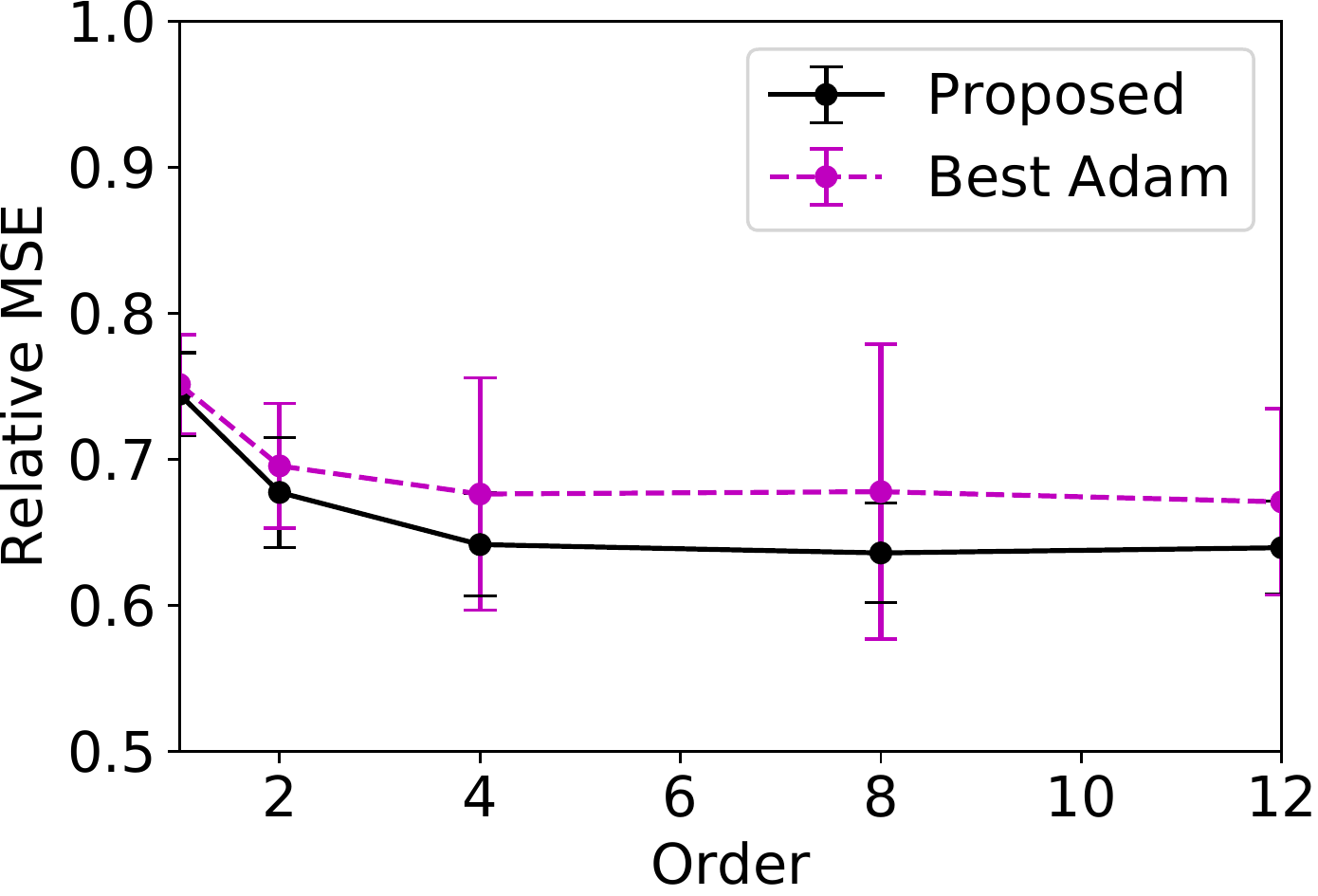}
  \includegraphics[width=0.33\linewidth]{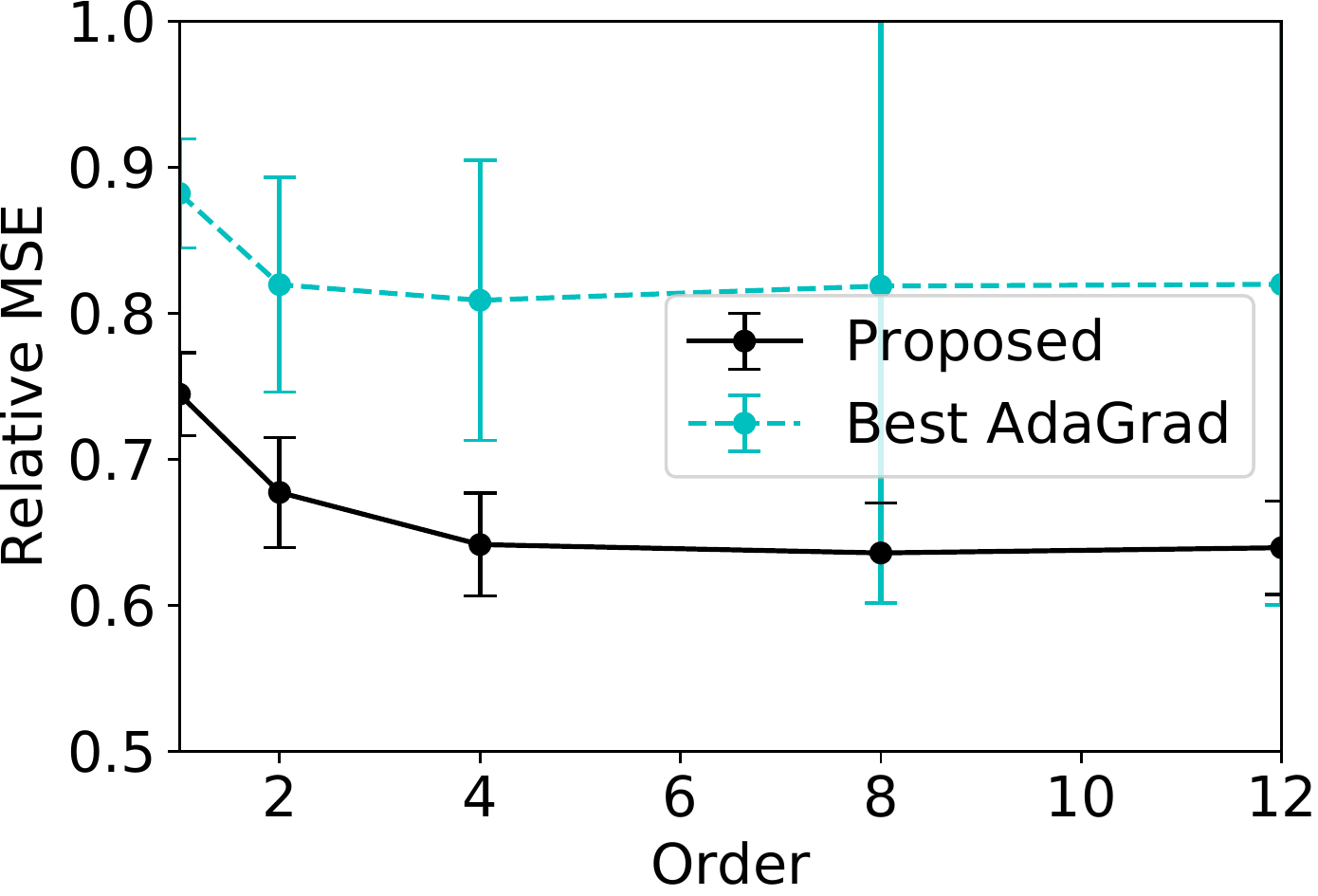}
  \includegraphics[width=0.33\linewidth]{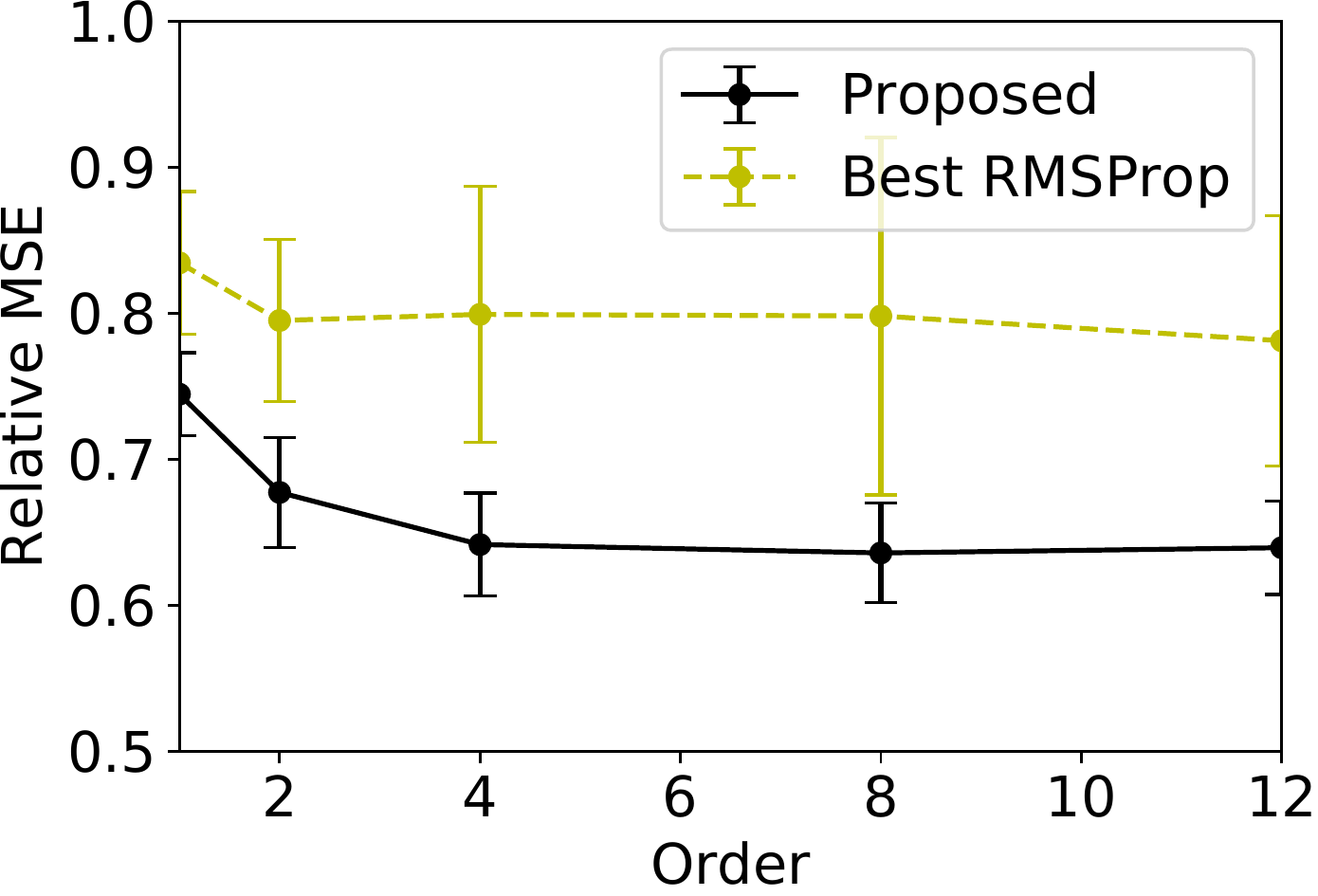}
  \caption{Details of the results on SSEC shown in Figure~\ref{fig:all}.
    Each panel shows the relative MSE of Algorithm~\ref{alg:adaptive}
    and a baseline (as indicated in the legend).  The error bars are
    drawn on the basis of the standard deviation of the MSE on each of
    the 10 intervals of equal length.}
  \label{fig:ssec}
\end{figure*}

Figure~\ref{fig:spx}-\ref{fig:ssec} shows the results in
Figure~\ref{fig:all} with error bars.  In each figure, the relative
MSE of Algorithm~\ref{alg:adaptive} and baselines are compared on a
particular financial index.  The baselines are vSGD, HGD, Almeida, Cogra, Adam, AdaGrad, and RMSProp, and each
panel shows the relative MSE with one of the baselines.  The figures
also show error bars, which are computed on the basis of the standard
deviation of the MSE on each of the 10 intervals of equal length.

The results with HGD, Almeida, and Cogra look similar to each other in
the figure.  This is because, for the financial time-series under
consideration, the prediction by these three methods was quite close
to the naive prediction that the absolute daily return stays unchanged
from the previous day.  Because we compute the error bars on the basis
of the standard deviation of the MSE on each of the 10 intervals of
equal length, the error bars of these three methods also look similar
to each other.

\clearpage

\clearpage

\clearpage

\clearpage

\bibliographystyle{abbrvnat}
\bibliography{app}

\end{document}